\documentclass{article}

\usepackage{arxiv}

\usepackage{booktabs}
\usepackage{graphicx}

\usepackage{amsmath}
\usepackage{amsfonts}
\usepackage{amssymb}

\usepackage[utf8]{inputenc} 
\usepackage[T1]{fontenc}    
\usepackage{hyperref}       
\usepackage{url}            
\usepackage{booktabs}       
\usepackage{amsfonts}       
\usepackage{nicefrac}       
\usepackage{microtype}      
\usepackage{lipsum}
\usepackage{graphicx}
\graphicspath{ {./images/} }

\usepackage{tcolorbox}
\usepackage{caption}
\usepackage{adjustbox}
\tcbuselibrary{skins}
\usepackage{geometry}
\usepackage{array}

\title{Clustering Algorithms and RAG Enhancing Semi-Supervised Text Classification with Large LLMs}

\author{
  Shan Zhong \\
  \texttt{zhongshanyes@gmail.com} \\
  \And
  Jiahao Zeng \\
  \texttt{jiahaozeng@link.cuhk.edu.cn} \\
  \AND
  Yongxin Yu \\
  \texttt{yuyongxin@cib.com.cn} \\
  \And
  Bohong Lin \\
  \texttt{linbohong@cib.com.cn} \\
}

\begin{document}
\maketitle
\begin{abstract}

This paper proposes a Clustering, Labeling, then Augmenting framework that significantly enhances performance in Semi-Supervised Text Classification (SSTC) tasks, effectively addressing the challenge of vast datasets with limited labeled examples. Unlike traditional SSTC approaches that rely on a predefined small set of labeled data to generate pseudo-labels for the unlabeled data, this framework innovatively employs clustering to select representative "landmarks" for labeling. These landmarks subsequently act as intermediaries in an ensemble of augmentation techniques, including Retrieval-Augmented Generation (RAG), Large Language Model (LLMs)-based rewriting, and synonym substitution, to generate synthetic labeled data without making pseudo-labels for the unlabeled data. Empirical results show that even in complex text document classification scenarios involving over 100 categories, our method achieves state-of-the-art accuracies of 95.41\% on the Reuters dataset and 82.43\% on the Web of Science dataset. Our approach significantly reduces the reliance on human labeling efforts and the associated expenses, while simultaneously ensuring high data quality and minimizing privacy risks. The finetuning results further show the efficiency of fine-tuning LLMs for text classification tasks, highlighting a robust solution for leveraging limited labeled data.

\end{abstract}


\section{Introduction}

In today's digital age, industries such as finance and banking process and collect vast amounts of data on a daily basis. However, a significant portion of this data remains unlabeled, and much of it cannot be utilized due to stringent privacy regulations. The difficulty in obtaining labeled data is multifaceted. Firstly, the cost associated with human labeling can be prohibitively high, especially when dealing with a large volume of similar or redundant samples. Secondly, concerns over privacy and confidentiality add an additional layer of complexity. The sensitive nature of the information often restricts or prohibits the direct use of substantial volumes of data from production environments. To tackle the issue of insufficient labeled data, Semi-Supervised Text Classification (SSTC) seeks to enhance classification accuracy by leveraging a small set of labeled data in conjunction with a larger set of unlabeled data. However, Traditional SSTC methods\cite{64ae66d03fda6d7f06848048}\cite{63e4cf6290e50fcafdb05e7c} often entail complex model architectures that are challenging to train and deploy. Moreover, these methods frequently rely on pseudo-labeling for unlabeled data. Yet, generating pseudo labels does not address privacy concerns since the original data remains unchanged. Additionally, errors introduced during the pseudo-labeling process can propagate through the system, potentially undermining the accuracy and reliability of the classification models. To effectively address these issues, this study introduces an innovative framework of fine-tuning LLMs for SSTC tasks by integrating a series of distinct components into a cohesive pipeline. Initially, clustering algorithms are employed to carefully select representative data points that effectively capture the diversity of the dataset while avoiding redundancy. Following this selection process, human experts manually label these carefully chosen examples, serving as high-quality retrieval data for subsequent processes. The labeled data then undergo a series of data augmentation techniques including word replacement, Large Language Model (LLMs)-based rewriting, and Retrieval-Augmented Generation (RAG) augmentation, to produce synthetic yet contextually relevant and accurately labeled training data.\\

In recent years, LLMs have achieved remarkable progress in the domain of artificial intelligence, functioning not only as generative models but also as probabilistic models with the capacity to output choice probabilities and make decisions. This can be achieved, for example, by assigning choice IDs such as A and B and calculating the logit of the last output word before the EOS token. Recent research indicates that large language models are excellent zero-shot and few-shot learners by incorporating examples and instructions into the prompt.\cite{5ed0e04291e011915d9e43ee}. Technologies such as Chain of Thought (CoT)\cite{61f753205aee126c0f9c20e3} and Tree of Thoughts\cite{6466fafbd68f896efaeb751d} have subsequently further demonstrated significant enhancements in multiple text classification\cite{624bb3a65aee126c0fea533e}\cite{61e781575244ab9dcbf999ab} benchmark tests by enabling LLMs to output reasoning before the answer, or by conducting further ensembles, i.e self-consistency\cite{62393e845aee126c0f1265e1} among multiple answers. Thus, employing LLMs as models for SSTC tasks can benefit from their generalization and texture understanding ability. Moreover, the popularity of LLMs, along with numerous developed training and deployment frameworks, makes them more attractive contrast to traditional SSTC models.\\

Beyond LLMs's proficiency in SSTC tasks, they also exhibit a remarkable capability to generate coherent text for data augmentation purpose. However, leveraging LLMs for data augmentation presents certain challenges, including a deficiency in domain-specific knowledge and the propensity for generating inaccurate or inconsistent content, often referred to as "hallucinations". Studies such as those conducted by \cite{6204827f5aee126c0f77dc55} found that the data generated using LLM alone may have poor correlation with the original data, especially for sequence pair tasks requiring subtle semantic relationships. \cite{63fd715990e50fcafd146bdb} further elucidate that LLM-generated data, when solely reliant on character-level or word-level inputs, can suffer from issues like low consistency and insufficient diversity, negatively impacting the performance of the fine-tuned model. Retrieval Augmented Generation (RAG)\cite{5ecce8d991e0119170395aab}, introduced in 2020, has become popular for its ability to link answers with professional knowledge without additional training, thus improving answer quality and reducing hallucinations. This is achieved by first retrieving relevant information from external knowledge, mainly through embedding models, and subsequently combining the retrieved information into the input prompt of LLMs to generate answers. For instance, \cite{6641743701d2a3fbfce99171} demonstrated that combining RAG with LLMs could significantly reduce the incidence of "hallucinations," improving text generation quality in specialized domains like law and medicine, where accuracy is paramount. Similarly, \cite{65a75ad1939a5f408261b5ae} showed that RAG could mitigate model illusions and biases in specific contexts such as agriculture. \\

However, methods like RAG or CoT that do not involve fine-tuning may sometimes produce outputs that do not fully align with the instructions given in the prompt. Additionally, appending content to the prompt and generating reasoning before providing an answer can lead to higher computational costs. In production environments that handle large volumes of data and require quick responses, a smaller, faster, and specialized model that can directly output the answer is often more desirable. Fine-tuning allows for the adjustment of a language model's behavior to conform to these desirable writing styles\cite{65824f86939a5f4082a849dd}. While \cite{6356022390e50fcafd3368cb} reveals that LLMs can improve themselves through the utilization of CoT prompting and self-consistency to generate high-quality answers from unlabeled data and fine-tuning based on these generated answers. Studies such as \cite{63a1750d90e50fcafd1f3b08} have further demonstrated that fine-tuning a model with fewer parameters through knowledge distillation from the CoT outputs of larger LLMs results in improved task performance on datasets like GSM8K. In this research, we developed a semi-supervised fine-tuning framework to enhance the text classification capabilities of smaller LLMs. Extensive experiments demonstrated that after undergoing our fine-tuning process, these smaller models outperformed their much larger, unfine-tuned counterparts.\\

Given that real-world datasets frequently contain duplicates, mislabeled instances, or lack labels altogether, and considering the substantial cost associated with obtaining expert annotations, it is essential to identify the most representative data points that accurately capture the dataset's characteristics for effective model fine-tuning. The study by \cite{5ce2d121ced107d4c63f4f73} explored the selection of representative days for power system investment planning, employing clustering techniques on dimensionality-reduced numeric time series data. Another work by \cite{60641d2d9e795e72406b6636} addressed the challenge of selecting representative subsets from big data streams, focusing on streaming algorithms that require minimal passes through the dataset while developing a metric to assess representativeness. Moreover, \cite{5d19db4d3a55ac7eb1cdf609} used smaller sized deep learning model in selecting representatives. Additionally, research by \cite{5d8898013a55acdc5ca088f4} has highlighted concerns related to sample selection bias, such as the repetition of redundant samples, the inclusion of outliers, or an overrepresentation from certain classes. Different from these works, our research uniquely focuses on textual data, employing various clustering algorithms to uncover patterns that can serve as exemplars to guide LLMs in generating more sythentic data that can improve LLMs finetunning accuracy. By carefully selecting "landmark" data points through clustering, our rewrite and word replacement augmentations target the most representative elements of the dataset, preserving a distribution similar to the original. The Retrieval-Augmented Generation (RAG) mechanism leverages these landmarks as references during the augmentation process. Controlled experiments show that employing representative landmarks for fine-tuning, as opposed to random selection, results in enhanced model accuracy.\\

Our study introduces a novel method for enhancing text classification datasets by integrating samples with varying vocabulary overlap into the original dataset. This approach broadens the coverage of case types while reinforcing predominant scenarios, thereby increasing data diversity. We demonstrate significant accuracy improvements by combining multiple augmentation techniques in specific proportions. Specifically, the use of LLMs for rewriting and word replacement augmentation of representative data points strengthens the majority cases in text classification. Meanwhile, the RAG technique expands dataset diversity and richness by generating new data from unlabeled sources, guided by the small set of labeled, representative data. Although numerous studies have focused on each minor aspect of text classification, few have  provided detailed explanations of the various stages involved in these tasks, such as data selection, preprocessing, labeling, data augmentation, and iterative training with different augmented datasets. Conversely, our approach is simple to implement and understand, with each step grounded in open-source packages. This compatibility ensures that our method is more practical and accessible, unlike other techniques that may present difficulties in verification and execution.\\

\section{Literature Review}

Traditional text data augmentation techniques typically involve operations such as randomly swapping, inserting, or deleting words, or using synonym replacements within a context, which can be achieved through word lists or by leveraging a trained language model \cite{5b1642d68fbcbf6e5a9b7f1a}. These methods have been applied to augment datasets in various domains, including legal documents \cite{62ac38835aee126c0f45cb08} and emails \cite{5ecbc8929fced0a24b5200c4}. Additionally, specialized augmentation techniques, such as simulating misspelled words and OCR errors, graph-structured augmentation, style transfer augmentation, and adversarial augmentation, are utilized to address particular challenges \cite{60ffe0c05244ab9dcb2cd725}. When it comes to task-specific augmentation that requires domain expertise or involves modifying documents with specific styles, especially when dealing with semi-structured texts where certain words must not be altered, a more sophisticated approach is essential \cite{62ac38835aee126c0f45cb08}. With the emergence of LLMs, researchers have started utilizing models like ChatGPT for generating and augmenting data. For example, one study \cite{63fd715990e50fcafd146bdb} used ChatGPT to produce diverse versions of clinical sentences while maintaining their original labels, subsequently validating and filtering these with a BERT model to ensure the quality of the augmented data. This method was primarily applied to short sentences. Since LLMs gained the capability to generate synthetic data, individuals have developed various strategies to filter high-quality data based on their needs. An illustrative case is a large reward model \cite{6670f8b601d2a3fbfc51511d} designed to evaluate data across a five-dimensional score encompassing Helpfulness, Correctness, Coherence, Complexity, and Verbosity.\\

Retrieval-Augmented Generation (RAG) \cite{5ecce8d991e0119170395aab} is a method aimed at enhancing generation performance by fetching text information from external knowledge bases and incorporating this information into the prompt to produce output. Since the advent of RAG, various studies \cite{65824f86939a5f4082a849dd} have made diverse efforts to improve RAG outputs through methods such as query rewriting, document re-ranking, and post-retrieval processing. More sophisticated methods include incorporating evaluation modules to measure the appropriateness of employing RAG compared to direct text generation, or adopting distinct specialized workflows, such as exploiting knowledge graphs \cite{662b0afe01d2a3fbfc65fd7f} or navigating through specific hierarchical levels of documents for retrieval, depending on the given context or scenario. Other examples include sophisticated paragraph-segmenting methods that reduce irrelevant noise, transforming queries to broaden or refine search intent, and the utilization of metadata filters to narrow down results. Additionally, hybrid retrieval models, which combine embeddings for semantic understanding with traditional tools like Elasticsearch and keyword matching, offer complementary strengths when employed in ensemble, or facilitate the switching of search strategies when dealing with large datasets to conserve time. Re-ranking or using different LLMs to refine selections, fine-tuning the retriever or generator, and iterative retrieval processes, where successive queries refine the result set based on the initial feedback, further enhance the accuracy and relevance of information retrieval. Collectively, these strategies improve search performance by adapting to user needs. Focused to text classifcation tasks, \cite{6466fafbd68f896efaeb7595} utilized retrieval-enhanced LLMs to generate text classification datasetes under zero-shot condition. Moreover, \cite{64ffca703fda6d7f06cd8a6e} proposed a Retrieval-Augmented framework to alleviate poor generalization issues existed in text classification taskes.\\

The Chain of Thought (CoT) method, initially introduced by \cite{61f753205aee126c0f9c20e3}, involves incorporating a sequence of intermediary reasoning steps that precede the formulation of conclusions. CoT prompting has shown a significant boost in accuracy across a wide range of different tasks. \cite{621c3d215aee126c0fe7e1de} provided a theoretical perspective on why prompting works. They experimented by replacing the correct labels with incorrect ones for the few-shot examples and found that this barely affected performance. Instead they found the distribution of the input text in the few-shot examples, the overall format of the few-shot examples, and the label space of the few-shot examples are the key factors contributing to performance. When employing CoT methodologies, the extraction of the final answer is typically achieved by parsing the generated text using regular expressions, which can be prone to extraction errors. \cite{664c009901d2a3fbfcde8518} developed a 500 million parameter language model (LLM) specifically designed to extract answers from the generated responses, demonstrating improved accuracy over regular expressions. Further exploration by \cite{634e194890e50fcafd24f6a2} revealed that CoT provides 10\% to 20\% accuracy gains over standard prompting in most tasks, except for those requiring external world knowledge, where its accuracy lags behind standard prompting.\\

Despite the abundance of technical reports on various models that open-source their model weights, there are relatively less publicly available reports as well datasets on the techniques for training and fine-tuning language models. The training of LLMs typically involves stages such as pre-training, instruction fine-tuning, and preference fine-tuning\cite{6181fdcc5244ab9dcb7a6715}\cite{647572e0d68f896efa7b79a5}. Our focus lies on the instruction fine-tuning aspects. \cite{63520de890e50fcafd60f4dd} investigated scaling the number of tasks, the number of data points, and the number of parameters. They found that multi-task instruction fine-tuning enhanced prediction accuracy: as the number of fine-tuned tasks increased, the prediction accuracy of an 8B parameter LLMs on the renowned Massive Multi-task Language Understanding (MMLU) dataset rose from 24.3\% to 26.3\%, 37.7\%, 47.7\%, and 49.3\% for 9, 89, 282, and 1836 fine-tuned tasks respectively. This margin diminishes as the size of the model parameter becomes larger. \cite{64225b7690e50fcafde12339} discussed about the dataset size with respect to the instruction fine-tuning performance. They selected instruction data of different scales from 200,000, 600,000, 1 million, to 2 million for comparison, and observed a steady overall improvement across different tasks as the trainable data increased. However, their training data was generated from ChatGPT with less than 200 high-quality human-labeled data for each task.\cite{646aeca9d68f896efa05a53c} demonstrated that LLMs can effectively learn interaction styles or formats through fine-tuning with fewer than 1,000 carefully annotated training prompts and responses, without the need for data augmentation or other tricks. \cite{6405a5bee368bb303e8c93ab} discussed about parameter efficient fine-tuning approaches and compared the accuracy among different tunning methods such as vanilla fine-tuning, prompt-tuning, prefix-tuning, LoRA and adapter for over 100 tasks. They found that vanilla fine-tuning have the highest average results while prefix tuning, LoRA and adapter are not far behind. They also state that they did not observe an increase in performance with the increase of the efficient tuning parameters for at least small-sized LLMs.\\


Semi-Supervised Text Classification (SSTC) tasks can benefit from Latent Dirichlet Allocation (LDA) topic models, which are probabilistic models optimized for parameters that effectively capture the thematic structure of unlabeled text instances. \cite{599c7ab8601a182cd26d943a} employed a self-training method beginning with an initially small labeled dataset and, for each new incoming document, demonstrated superior performance compared to traditional Term Frequency-Inverse Document Frequency (TF-IDF) representation methods and supervised classifiers. Conversely, \cite{55a6bbb065ce054aad733bcc} suggested a technique focusing on the set of non-examples that do not fall into any of the categories of interest. This approach can be used to approximate the prior distribution of the classification function, thus improving the accuracy of methods like AdaBoost in SSTC tasks. Furthermore, \cite{600fe792d4150a363c2271eb} introduced an iterative self-pretraining method that uses two classifiers: one labels a subset of unlabeled data, while the other trains on both labeled and pseudo-labeled data, filtering out unreliable labels using confidence thresholds. Final classification is achieved through ensemble predictions from both classifiers.\\

\section{Methods and Datasets}
	
\definecolor{titleColor}{RGB}{0,39,76} 
\definecolor{frameColor}{RGB}{100,149,237} 
\definecolor{bgColor}{RGB}{230,242,255} 

\newcommand{\thickarrow}{\ensuremath{\mathbf{\Longrightarrow}}}

\begin{figure}[h]
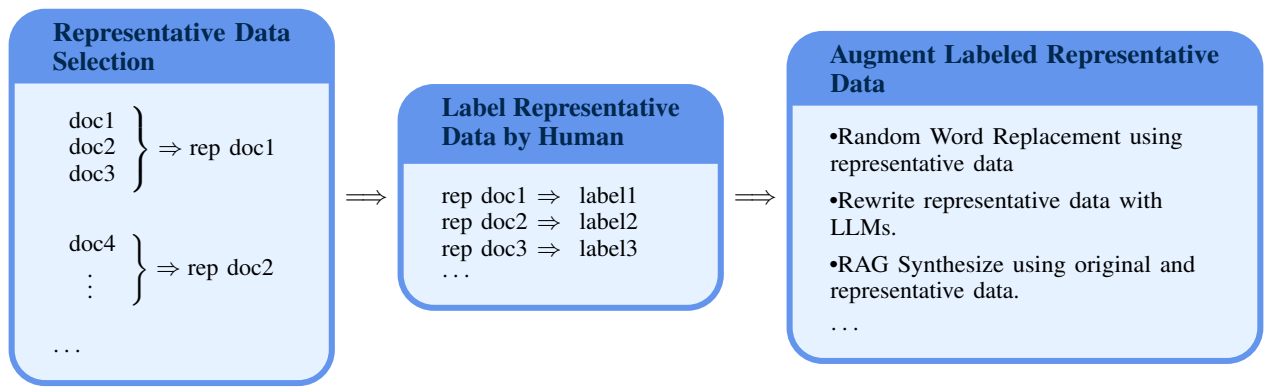

\noindent
\begin{minipage}[c]{0.26\linewidth}
    \begin{tcolorbox}[
        colback=bgColor,
        colframe=frameColor,
        fonttitle=\bfseries\color{titleColor},
        title=Representative Data Selection,
        halign=left,
        arc=4mm,
        boxrule=0.8mm, 
        colupper=black 
    ]
        \begin{flushleft}\footnotesize
            $\displaystyle
            \left.
            \begin{array}{l}
                \text{doc1}\\
                \text{doc2}\\
                \text{doc3}
            \end{array}
            \right\}
            \Rightarrow \text{rep doc1}
            $
            \vspace{0.5cm}

            $\displaystyle
            \left.
            \begin{array}{l}
                \text{doc4}\\
                \ \ \vdots
            \end{array}
            \right\}
            \Rightarrow \text{rep doc2}
            $
            \vspace{0.5cm}

            $\cdots$
        \end{flushleft}
    \end{tcolorbox}
\end{minipage}%
\hfill \thickarrow \hfill 
\begin{minipage}[c]{0.26\linewidth}
    \begin{tcolorbox}[
        colback=bgColor,
        colframe=frameColor,
        fonttitle=\bfseries\color{titleColor},
        title=Label Representative Data by Human,
        halign=left,
        arc=4mm,
        boxrule=0.8mm, 
        colupper=black 
    ]
        \footnotesize
        rep doc1 $\Rightarrow $ \  label1\\
        rep doc2 $\Rightarrow $ \  label2\\
        rep doc3 $\Rightarrow $ \  label3\\
        $\cdots$
    \end{tcolorbox}
\end{minipage}%
\hfill \thickarrow \hfill
\begin{minipage}[c]{0.38\linewidth}
    \begin{tcolorbox}[
        colback=bgColor,
        colframe=frameColor,
        fonttitle=\bfseries\color{titleColor},
        title=Augment Labeled Representative Data,
        halign=left,
        arc=4mm,
        boxrule=0.8mm, 
        colupper=black 
    ]
        \footnotesize
        \textbullet Random Word Replacement using representative data\\[4pt]
        \textbullet Rewrite representative data with LLMs.\\[4pt]
        \textbullet RAG Synthesize using original and representative data.\\[4pt]        
        $\cdots$
    \end{tcolorbox}
\end{minipage}%
\caption{Synthetic Data Generation Workflow for Fine-Tuning Classification LLMs}
\label{fig:workflow}
\end{figure}
	
As Figure\ref{fig:workflow} shown, our methodology unfolds in three distinct phases: In the initial phase, the emphasis lies on judiciously selecting a limited number of instances from a texture dataset for manual annotation. The dataset, denoted as \(X = \{doc_1, doc_2, \ldots, doc_n\}\), comprises individual documents where each \(doc_i\) symbolizes a sample from the original documents. The objective of this phase is to identify the most effective sampling strategy, \(\pi\), under the practical constraint that only a subset \(X_{\pi}^{m}\) of \(m\) samples can be annotated due to the inherent limitations of human annotation resources. This subset \(X_{\pi}^{m}\) of \(m\), drawn from the dataset \(X\) using the selection method \(\pi\), is subsequently assessed by an evaluation metric \(Q\), which we elect to be the several metrics that measure the performance of clustering using \(X_{\pi}^{m}\). Desbribed in a mathematical way\cite{65c588f2939a5f4082298ab2}, the quest revolves around finding the best selection approach, \(\pi^*=\arg\max_{\pi} Q\left(X_{\pi}^{m}, \pi\right)\), which optimizes the chosen performance of metric \(Q\). To attain our specific objective, we utilized unsupervised clustering approaches to identify a greater quantity of smaller and more homogeneous clusters of documents, given the lack of a predefined target distribution. This method is more favorable for identifying representative landmark data that reflect the diversity of the entire dataset, rather than forming a few large clusters consistsing broader categories with a large number of documents. By forming these smaller, more concentrated clusters, we then utilize the center of each cluster as our chosen landmarks \(X^{m} =\pi^*(X)\).\\

Building on these carefully selected landmarks, we proceed to the annotation phase. During this process, experts not only follow predefined guidelines but also develop their own nuanced standards through iterative refinement. This dynamic approach enhances annotation quality and consistency, ensuring that the synthetic data more accurately reflects real-world scenarios. We denote the representative landmarks as \((X^{m}, Y^m)\), where \(Y^m\) are annotated by human experts after obtaining \(X^{m}\). With these high-quality labeled landmarks in place, we transition into the data augmentation phase. Here, our objective is to expand the labeled dataset through a variety of augmentation strategies. These strategies include leveraging WordNet packages for word substitution, employing LLMs to rewrite documents, and generating data via RAG. In cases of random replacement using WordNet, a method similar to the one described in \cite{5ecbc8929fced0a24b5200c4} is adopted. Specifically, the WordNet \cite{56d84e6cdabfae2eeef002eb} lexical database is utilized to identify a set number of the closest synonyms, which are then randomly replaced with each other. For LLM-based rewriting augmentation, we follow the prompt strategy detailed in \hyperlink{tcb:rewrite1}{Table C.3}. For data augmentation involving WordNet substitutions and LLMs rewriting, only the \(X_{\pi}^{m}\) elements undergo augmentation, whereas the corresponding labels \(Y^m\) remain unaltered.\\

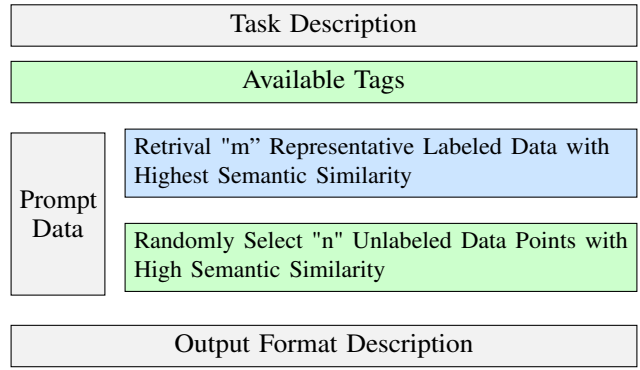
\begin{figure}[h]
\centering
\begin{tikzpicture}[node distance=1cm and 1cm]
    \definecolor{lightgrey}{RGB}{242,242,242}
    \definecolor{lightblue}{RGB}{204,229,255}
    \definecolor{lightgreen}{RGB}{204,255,204}
    \definecolor{orange}{RGB}{255,153,0}
    \node[fill=lightgrey, draw, text width=8cm, align=center] (task) at (0,2.5) {Task Description};
    \node[fill=lightgreen, draw, text width=8cm, align=center] (labels) at (0,1.75) {Available Tags};
    \node[fill=lightblue, draw, text width=6.5cm, align=left] (ref1) at (0.75,0.675) {{\small Retrival "m” Representative Labeled Data with Highest Semantic Similarity}};
    \node[fill=lightgreen, draw, text width=6.5cm, align=left] (ref2) at (0.75,-0.575) { {\small Randomly Select "n" Unlabeled Data Points with High Semantic Similarity}};
    \node[fill=lightgrey, draw, text width=1cm, align=center, minimum height=2.15cm] (refdata) at (-3.5,0) {Prompt Data};
    \node[fill=lightgrey, draw, text width=8cm, align=center] (output) at (0,-1.75) {Output Format Description};
\end{tikzpicture}
\caption{RAG Augmentation template for Fine-Tuning Classification LLMs}
\label{fig:ragworkflow}
\end{figure}

In our approach to RAG-based document generation, we integrate both labeled landmarks from similar neighborhood clusters and unlabeled samples within those same clusters, as illustrated in Figure~\ref{fig:ragworkflow} and summarized in \hyperlink{tcb:augment1}{Table C.4}. For each original document, we identify the five most likely clusters it may belong to and select labeled representatives from these clusters. Next, we randomly pick three unlabeled samples from the same cluster, merge them with the five chosen labeled landmarks from neighboring clusters, and combine them with the original document to create a diverse prompt set. This method does not require including all documents within the same cluster; instead, randomly selecting reference documents enhances the diversity of the generated content.\\

In the concluding phase of our project, we leverage the fine-tuning capabilities of pre-trained LLMs to enhance their performance. Fine-tuning LLMs allows us to produce more consistent answers compared to CoT methods and enables smaller models by parameter count to achieve performance comparable to larger models. The prompts designed for training the LLMs are outlined in \hyperlink{tcb:generate1}{Table C.1}, while for label prediction tasks, the prompt structure is presented in  \hyperlink{tcb:generate2}{Table C.2}. During the training phase, the model is instructed to output labels in the `[label1, label2, ...]' format. An explicit stop condition is implemented so that the model starts generating text with an opening bracket `[' and stops immediately after outputting a closing bracket `]'. Furthermore, to guarantee that the model only produces desired responses, we tokenize the available labels and constrain the model to output only the token IDs associated with these labels. For single-labeled data, the format `[label]` can be utilized; for multi-labeled data, if the training data are organized by importance from highest to lowest, LLMs will learn to output labels in the sequence `[label1, label2, ...]` reflecting this hierarchy. In hierarchical labeling scenarios, the training data can be structured as `[tier1, tier2, ...]`, where `tier1` labels come from a predefined list of tier1 labels, and `tier2` labels follow accordingly, thus making the model's output highly adaptable.\\

The fine-tuning procedures were conducted within the Llamafactory framework \cite{65fb94f713fb2c6cf6830016}. For all experiments, a total of 80GB GPU memory was provided by two A100 GPUs. For tasks that involved augmentation using the LLMs rewrite and RAG methodologies—where no additional training is required—the Qwen2.5-72B model was utilized. Conversely, for the fine-tuning experiments that necessitated iterative training and parameter adjustments, the comparatively smaller Qwen2.5-0.5B model was selected to ensure efficient resource utilization.\\

To conduct experiments and compare our work with others, we utilized several publicly available text datasets: The WOS-46985\cite{5a260c8617c44a4ba8a31d58} dataset which is a subset of the Web of Science Dataset that comprises approximately 2.8 million scientific papers, the subset contains the abstract of 46,985 published paper, seven broad categories, including computer science, electrical engineering, psychology, mechanical engineering, civil engineering, medical sciences, and biochemistry, and 134 specific subfields. The charastric of published paper abstract made the length of the text longer than classification with just one sentence. The ApteMod version of the Reuters-21578\cite{53e9b63ab7602d9704193a62} corpus includes 10,788 documents from Reuters financial newswire service dating back to 1987, grouped into 90 distinct topics such as corn, crude oil, earnings, gold, interest, and trade. Each document is labeled with one or more categories. The corpus is commonly used for training and testing machine learning models, particularly for multi-label text classification tasks.\\

The dataset is structured into three distinct subsets: 20\% for testing, 30\% for validation, and 50\% for training. Within the 50\% training subset, the true labels are hidden, but full access to the text context is provided for clustering, retrieval, and data augmentation.  It is important to note that the 30\% validation and 20\% testing sets are strictly reserved for evaluating the performance of the models and not be used for data augmentation or clustering. This ensures an unbiased assessment of our procedure's capabilities.
 
\begin{table}[h!]
    \centering 
    \caption{Comparison of Text Datasets}
    \label{tab:my_label}
    \begin{tabular}{@{}lll@{}} 
        \toprule
        Dataset            & WOS-46985 & Reuters-21578 \\
        \midrule
        Dataset Size       & 46,985     & 10,788         \\
        Category Number    & 134       & 90            \\
        Average word count & 199.8     & 164.9         \\
        Train Size             &    23,492       & 5,394         \\
        Validation Size         & 14,095          & 3,883         \\
        Test Size               &     9,398      & 2,158         \\
        \bottomrule
    \end{tabular}
\end{table}

\section{Metric}
\subsection{Homogeneity}

The Homogeneity\cite{53e9ab5ab7602d97034f000f} metric assesses whether each cluster contains data points exclusively from a single class, reflecting the purity of the clustering relative to a ground truth. Mathematically, homogeneity is defined as:

\[ H(L_{true}, L_{pred}) = 1 - \frac{H(L_{pred} \mid L_{true})}{H(L_{true})} \]

where \( H(L_{pred} \mid L_{true}) \) is the conditional entropy and \( H(L_{true}) \) is the entropy of the true labels. This metric is invariant under label permutations and non-symmetric; swapping true and predicted labels yields the completeness score. A score of 1 signifies perfect homogeneity.

\subsection{Silhouette Score}

The Silhouette Score, \( s(i) \), measures how much an object \( i \) is closer to its own cluster than to others. It is defined as:

\[
s(i) = \frac{b(i) - a(i)}{\max(a(i), b(i))}
\]

Here, \( a(i) \) is the mean distance between \( i \) and all other points in its cluster, and \( b(i) \) is the smallest mean distance between \( i \) and points in any other cluster. Scores range from -1 to 1, with higher values indicating better cluster matches and values near 0 suggesting ambiguity.

\subsection{Jaccard Similarity}

The Jaccard Similarity is a measure that indicates the closeness of two sets by comparing the size of their intersection to the size of their union. It is defined as:

\[
J(A, B) = \frac{|A \cap B|}{|A \cup B|}
\]

Where \( |A \cap B| \) is the number of elements in the intersection of sets \( A \) and \( B \), and \( |A \cup B| \) is the number of elements in their union. The Jaccard Similarity coefficient ranges from 0 to 1, where a score of 1 means the sets are identical, and 0 indicates no overlap.

\section{Experiment}

In the initial phase of our experiment, we applied a variety of clustering algorithms—including Bisecting K-Means, Balanced Iterative Reducing and Clustering using Hierarchies (BIRCH), Hierarchical Clustering, and Gaussian Mixture Models—to both the Reuters and WOS datasets. This was also done to compare the efficacy of embedding techniques against traditional TF-IDF approaches. To validate these methodologies, we used the validation set that included known labels, allowing us to calculate metrics such as Homogeneity, detailed in Appendix \ref{appendix:cluster}. Following these evaluations, we observed that embedding methods perform better than TF-IDF, and that the Gaussian Mixture Model performs better when the number of clusters is relatively small. Conversely, for a larger number of clusters—such as 1200 or more—Hierarchical Clustering yields superior results. We selected the Gaussian Mixture Model for further analysis because we chose a smaller number of clusters. For feature representation, we employed the BGE-m3\cite{65c19b55939a5f40825f9ad2}  Embedding method, which can embed up to 8,192 tokens into a 1,024-dimensional vector.\\

\begin{figure}[h]
\centering
\includegraphics[width=1\textwidth]{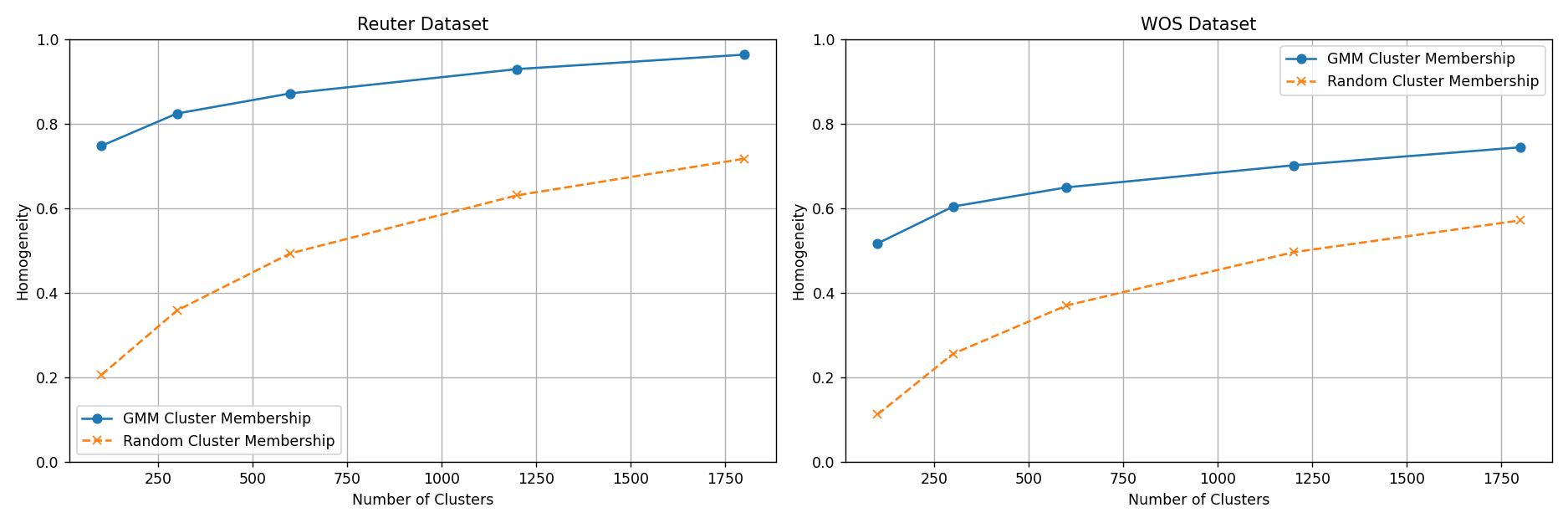}
\caption{Homogeneity against cluster size for the Reuters-21578 and WOS-46985 validation datasets, where clusters are assigned by Gaussian Mixture Models (GMM), with homogeneity calculated using true labels.}
\label{fig:clusters_homo}
\end{figure}

\begin{figure}[h]
\centering
\includegraphics[width=0.5\textwidth]{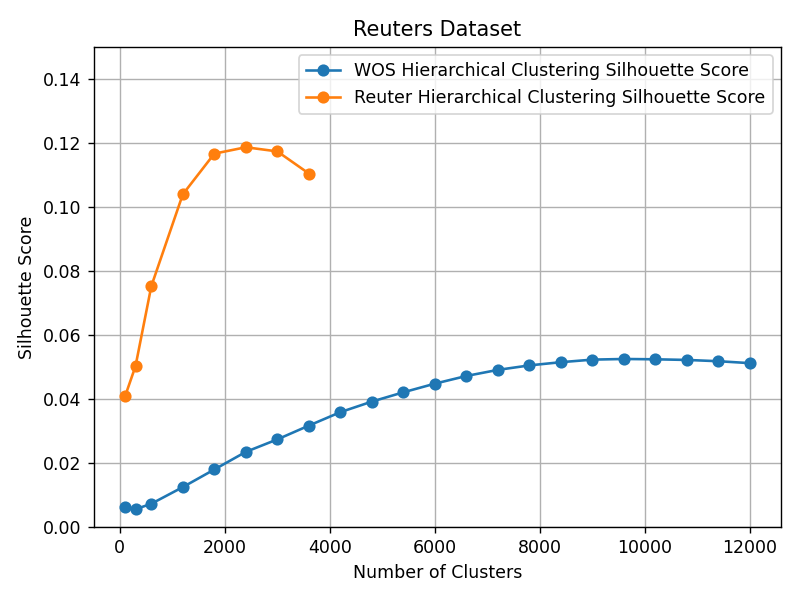}
\caption{Silhouette score against cluster size for the Reuters-21578 and WOS-46985 training datasets, where clusters are assigned by Hierarchical Clustering. Calculating the Silhouette score does not require true labels.}
\label{fig:clusters_silhouette}
\end{figure}

As illustrated in Figure~\ref{fig:clusters_homo}, Homogeneity increases with the number of clusters. It is evident that classifying the WOS dataset poses a greater challenge compared to the Reuters dataset, as indicated by the comparatively lower level of homogeneity achieved by WOS.  Additionally, it is observed that randomly assigned cluster membership gradually closes the gap with model-assigned memberships as the number of clusters increases. In scenarios where we encounter a completely new dataset without access to true labels, the silhouette score becomes an valuable metric for assessing clustering quality, as Figure~\ref{fig:clusters_silhouette} shown. In the Silhouette Score graphs, we see the score begins to decline when the number of clusters exceeds a threshold. For Reuters, this occurs at about 1,800 clusters (33.3\% of training data), and for WOS, at around 10,200 clusters (43.4\% of training data). However, given that the objective of our clustering process is to identify landmark points for labeling, the number of clusters we can manage is constrained by the availability of human resources for labeling. For the Reuters dataset, we selected 300 clusters, whereas for the WOS dataset, we chose 600 clusters.\\

\begin{figure}[h]
\centering
\includegraphics[width=1\textwidth]{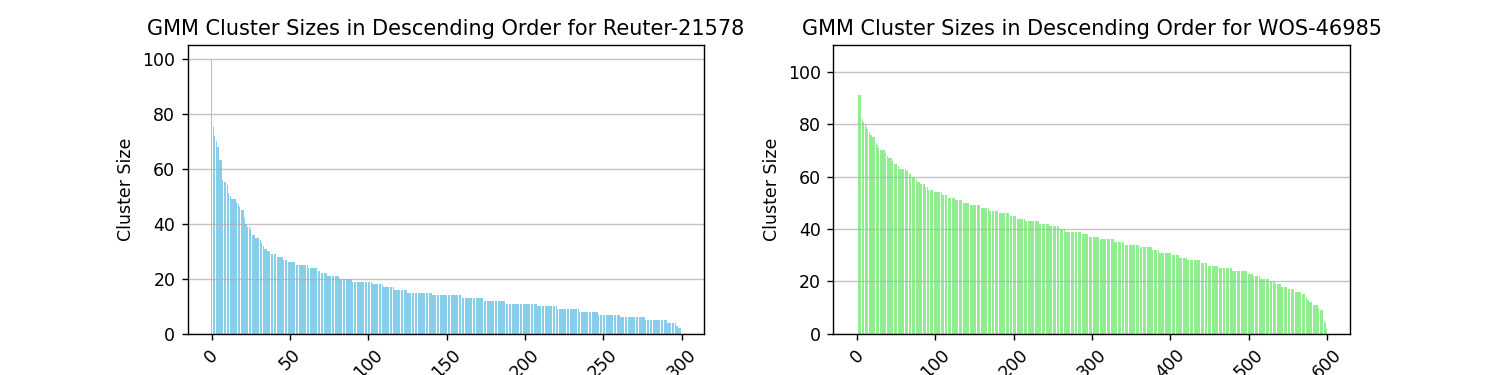}
\caption{GMM Cluster Size Distribution for Reuters-21578 and WOS-46985 Datasets}
\label{fig:clusters}
\end{figure}

We then proceed to analyze the training dataset, as illustrated in Figure \ref{fig:clusters}. The resulting clusters exhibit a distribution that is somewhat less skewed than anticipated. Typically, clusters are unevenly distributed, with a majority of elements concentrated within a few large clusters, while many smaller clusters contain only one or two members. For the Reuters dataset, which comprises a total of 300 clusters, the largest cluster contains 100 elements, while the smallest includes just 3 members. The median size of the clusters is 14, and the 95th percentile is at 5 members. The silhouette score for this dataset is 0.0472. For the WOS dataset, consisting of 600 clusters, the largest cluster has 105 elements, and the smallest has only 2 members. The median size of the clusters is 37, and the 95th percentile is at 15 members. The silhouette score for the WOS dataset is 0.0152.\\

Importantly, we assume we did not have access to the labels of the training set. Instead, we strategically selected a proportion of the training data to reveal true labels, identifying key data points as landmarks. To refine the selection of representative samples from each cluster, we implemented a multi-step process. Initially, all samples within each cluster were indexed. Subsequently, LLMs were employed to identify and select the sample that most accurately embodies the core characteristics of its respective cluster. This was achieved by instructing the LLMs to return the index corresponding to the most representative sample, as illustrated by prompts in \hyperlink{tcb:repchoose1}{Appendix C.5}. Notably, selecting the sample closest to the center as a representative landmark, calculated through embedding methods, is also an effective approach and the choice between methods can be guided by specific requirements.\\

Following the selection and revelation of the true labels for the most representative landmarks from each cluster, we proceeded to utilize all available unlabeled data, along with the labeled representatives, to generate additional training data. For RAG generation using the prompt attached in \hyperlink{tcb:augment1}{Appendix C.4}, we generated three unique samples for each document by varying the randomly selected reference documents. Subsequently, we applied regular expressions to parse out the relevant context and labels from the generated texts. Out of the 16,182 documents generated for the Reuters dataset, 15,473 successfully underwent the regular expression-based extraction of context and labels. Similarly, for the WoS dataset, 62,886 documents passed through the extraction process. Additionally, for the rewriting task using LLMs as prompt detailed in \hyperlink{tcb:rewrite1}{Appendix C.3}, we generated ten rewritten samples for each of the landmark documents within the datasets. This was conducted with the hyperparameter temperature set to 0.3, yielding a total of 3,000 rewritten documents for the Reuters dataset and 6,000 for the WoS dataset. Since these generated samples did not require label extraction, they all successfully passed the regular expression filtering phase. Finally, we applied a similar process using WordNet for random replacement, generating an additional 3,000 rewritten documents for the Reuters dataset and 6,000 for the WoS dataset.

\begin{table}[htbp]
\centering
\small 
\begin{tabular}{>{\raggedright\arraybackslash}p{2.5cm} c c c c}
\toprule
 & \multicolumn{2}{c}{Reuters} & \multicolumn{2}{c}{WOS} \\
\cmidrule(lr){2-3} \cmidrule(lr){4-5}
 & Jac. Sim Among Gen. & Emb. Sim Orig-Gen &  Jac. Sim Among Gen. & Emb. Sim Orig-Gen\\
\midrule
WordNet & 94.57\% & 97.75\% & 97.13\% & 98.52\% \\
LLM Rewrite & 78.36\% & 93.51\% & 76.32\% & 96.35\% \\
RAG Augment & 49.12\% & 92.17\% & 40.91\% & 90.84\% \\
\bottomrule
\end{tabular}
\caption{Comparison of different augmentation methods on Reuters and WOS datasets}
\label{tab:sim_comparison}
\end{table}

\begin{figure}[h]
\centering
\includegraphics[width=1\textwidth]{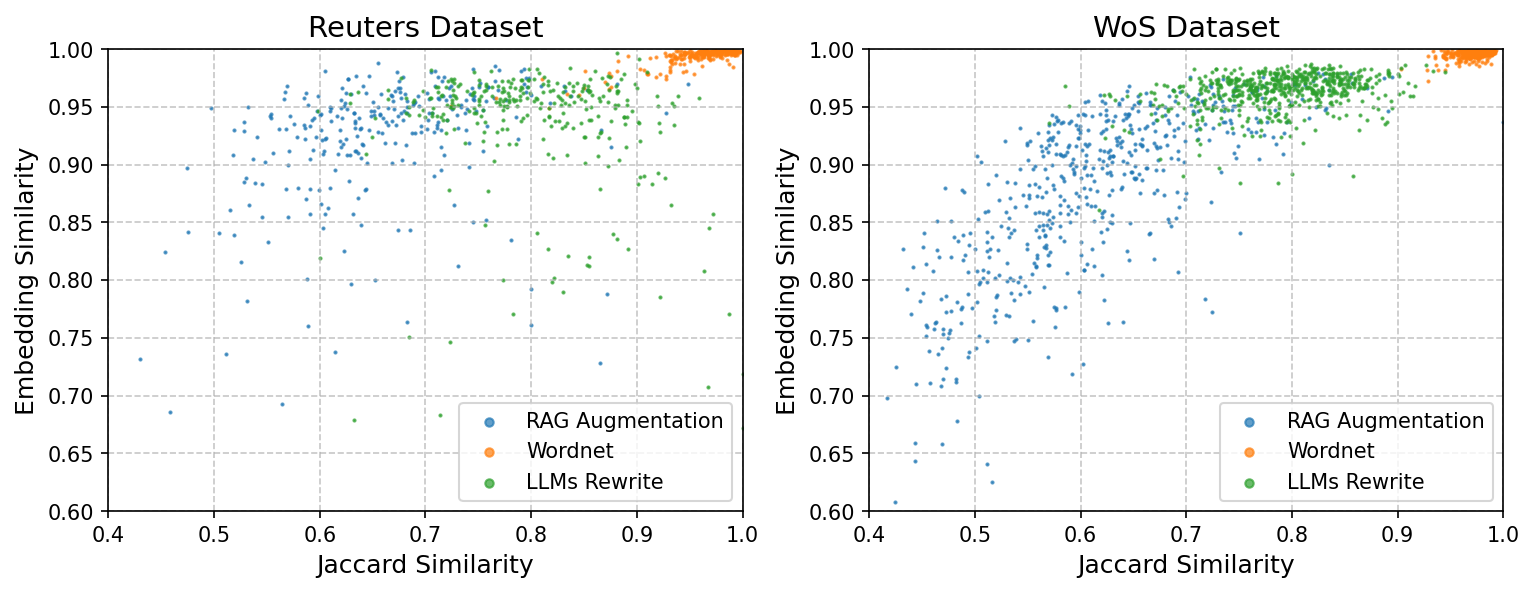}
\caption{Impact of different Text Augmentation method on Jaccard Similarity (Vocabulary Overlap) and Embedding Similarity (Sementic Similarity) among Generated Samples: The figure illustrates that from  traditional WordNet random synonym replacement to LLMs rewrite to RAG augmentation, the vocabulary overlap decrease, indicating the generation of more diverse samples.}
\label{fig:embd_jacc}
\end{figure}

To evaluate the impact of various text augmentation techniques on the Reuters and WOS datasets, we conducted a comprehensive analysis focusing on the similarity between the original and augmented documents, as well as the diversity among the augmented samples derived from the same source document. Two primary metrics were used to quantify these aspects: Jaccard Similarity, which measures the degree of vocabulary overlap between sets of words, and Embedding Similarity, which assesses semantic similarity. Our findings, summarized in Table \ref{tab:sim_comparison} and illustrated in Figure \ref{fig:embd_jacc}, show that different augmentation methods yield varying levels of vocabulary overlap and semantic similarity:

\begin{itemize}
    \item \textbf{WordNet Random Synonym Replacement:} This method produced the highest vocabulary overlap, with Jaccard Similarity scores of 94.57\% for the Reuters dataset and 97.13\% for the WOS dataset, calculated among the 10 samples generated from each landmark document, indicating that this technique preserves much of the original text's structure and vocabulary. 
    
    \item \textbf{LLMs Rewrite:} This method provided a balanced approach, achieving Jaccard Similarity scores of 78.36\% for the Reuters dataset and 76.32\% for the WOS dataset, calculated among the 10 samples generated from each landmark document. It effectively introduced new variations while maintaining the core essence of the original content.

    \item \textbf{RAG Data Augmentation:} This method resulted in the most diverse set of samples, with the lowest Jaccard Similarity scores of 49.12\% for the Reuters dataset and 40.91\% for the WOS dataset, calculated among the 3 samples generated using the RAG representative documents and other similar random components. The inclusion of varied, randomly chosen supporting documents notably increased content diversity.
\end{itemize}

From RAG augmentation to LLMs rewrite to traditional WordNet random synonym replacement, the vocabulary overlap decreases, indicating the generation of more diverse samples. Despite this lower vocabulary overlap, the embedding-based similarity remained high, with scores of 92.17\% for the Reuters dataset and 90.84\% for the WOS dataset. This suggests that while RAG augmentation introduces significant changes, it maintains the core meaning and semantic coherence of the original text. For LLMs rewrites and WordNet augmentation, they reinforce the representation of the majority cases, ensuring that the model performs robustly on common scenarios without being overly influenced by edge cases. We will further demonstrate that combining documents with varying similarity levels can significantly boost prediction accuracy. This strategy leverages the core information and semantic coherence preserved in high-similarity documents while adding new perspectives from lower-similarity ones. As a result, it enhances the model's learning and generalization, leading to more accurate predictions.\\

We then discuss in detail how labels were generated alongside RAG augmentation. For the Reuters dataset, we configured the LLMs to produce multiple labels for each document, with individual labels delineated by commas. In contrast, for the WoS dataset, the LLM was instructed to generate a hierarchical labeling structure, beginning with the domain and then specifying the sub-domain, with each level separated by commas. We also observed that there are cases where RAG does not perform well, such as with the "lei" category in the Reuters dataset. The "lei" category, representing the Leading Economic Index, saw no augmentation because the LLMs failed to recognize "lei" without sufficient context. A more detailed report about the distribution of the generated labels is provided in the Appendix~\ref{sec:gen_data_desc}.\\

\section{LLM Finetunning}

In our experiments, we utilized augmented data to train a Qwen2.5 0.5B language model to evaluate the performance of different augmented datasets. For the Reuters dataset, we evaluated the model using three metrics: \textbf{Part Match}, \textbf{All Match}, and \textbf{In Right Order}, based on a test set of \textbf{2,158} samples. These metrics are defined as follows:

\begin{itemize}
    \item \textbf{Part Match}: At least one of the generated tags aligns with one of the actual tags.
    \item \textbf{All Match}: All generated tags perfectly match the true tags.
    \item \textbf{In Right Order}: All tags match and are presented in the correct order of importance.
\end{itemize}

For the Web of Science (WoS) dataset, we evaluated the model using  \textbf{Domain Match} and \textbf{Area Match}, which measure the accuracy of predictions at different hierarchical levels. The test set size was \textbf{9,398} samples. These metrics are defined as follows:

\begin{itemize}
    \item \textbf{Domain Match}: The generated domain matched with the true domain.
    \item \textbf{Area Match}: In addition to the domain matching, the generated area also matches the true area.
\end{itemize}

We then investigate the performance of for these metrics under LLMs trained with different seperated datasets. This investigation aims to understand how these methods can enhance the quality and diversity of the training data, potentially leading to further improvements in model performance. The datasets used in this study are detailed below:

\begin{itemize}
    \item \textbf{Landmark Data}:  Selected landmarks believed to be indicative of the core themes within each dataset, 300 from Reuters and 600 from WoS. 
    \item \textbf{Landmark-WordNet Data}: The landmark data was augmented tenfold using WordNet, expanding the dataset to 3,000 samples from Reuters and 6,000 from WoS. 
    \item \textbf{CoT+RAG}: This method involves using a larger Qwen 2.5 72B model with RAG to provide more information and CoT prompt instructions to sequentially generate answers. The prompts used for this process are detailed in \hyperlink{tcb:labeling1}{Appendix C.6}, and no parameter tuning was involved.    
    \item \textbf{Rewrite Data}: In this approach, the Landmark Data is rewritten using a Qwen 2.5 72B language model, as detailed in \hyperlink{tcb:rewrite1}{Appendix C.3}. This expands the dataset to 3,000 samples from Reuters and 6,000 from Web of Science (WoS).
    \item \textbf{RAG Augmented Data}: In this approach, each document is augmented using a Qwen 2.5 72B language model, as detailed in \hyperlink{tcb:augment1}{Appendix C.4}. This expands the dataset to 15,473 samples from Reuters and 62,886 from WoS.
    \item \textbf{Combined Data}: This method combines the WordNet, LLM rewriting, and RAG augmentation techniques by merging their training data together. 
    \item \textbf{Original Data}: For comparison only, we also list the performance of the model when using all training data with their true labels. The original datasets contain \textbf{5,394} true labels for Reuters and \textbf{23,492} true labels for WoS.
\end{itemize}

\begin{table}[ht]
\centering
\begin{tabular}{llll}
\toprule
\textbf{Dataset} & \textbf{Part Match} & \textbf{All Match} & \textbf{In Right Order} \\ 
\midrule
Reuters, CoT+RAG & 82.52\% & 73.03\% & 70.60\% \\
Reuters, Landmark data & 90.85\% & 82.35\% & 75.86\% \\
Reuters, Landmark wordnet data & 93.05\%  & 83.97\% & 78.82\% \\
Reuters, Rewrite Data & 91.47\% & 83.73\% & 78.08\% \\
Reuters, RAG Augment Data & 95.14\% & 85.68\% & 78.87\% \\
Reuters, Combined Augment Data & \textbf{95.41\%} & \textbf{89.02\%} & \textbf{81.60\%} \\
Reuters, Original Data, for reference only & 96.66\% & 91.89\% & 85.77\% \\
\bottomrule
\end{tabular}
\caption{Detailed results for Reuters dataset}
\label{tab:reuters_detailed}
\end{table}

From the results presented in Table \ref{tab:reuters_detailed}, it is clear that even with a larger model like Qwen 2.5 72B employed for CoT+RAG, its performance does not match that of a much smaller, parameter fine-tuned model such as Qwen 2.5 0.5B, which outperforms the larger model across multiple metrics. Specifically, for the Reuters dataset, the RAG Augmented Data method significantly outperformed other data augmentation techniques. Notably, despite using only 300 landmark labels, the RAG Augmented Data achieved a part match rate that was only 1.52\% lower than the result obtained using the full set of 5,394 training data labels, which is a remarkable achievement. On the other hand, it is worth noting that when using only RAG augmented data, the accuracy for "All Match" is comparatively lower. This suggests that while RAG augmented data excel at partial matching, they may be more challenging to clearly differentiate for complete matches. Moreover, the Combined Augmentation method, which integrates WordNet, LLM rewriting, and RAG augmentation, achieved Part Match, All Match, and In Right Order scores of 95.41\%, 89.02\%, and 81.60\%, respectively, on the Reuters dataset. This demonstrates that our strategy of integrating data with varying levels of vocabulary overlap has indeed contributed to improved metrics. The ensemble approach effectively leverages the strengths of different data augmentation techniques, leading to a more robust and accurate model, particularly for the "All Match" and "In Right Order" categories, thereby addressing the limitations observed with RAG data.\\

\begin{table}[ht]
\centering
\begin{tabular}{lll}
\toprule
 & \textbf{Domain Match} & \textbf{Area Match} \\ 
\midrule
WOS, CoT+RAG & 63.54\% & 43.69\% \\
WOS, Landmark data & 73.85\% & 47.48\% \\
WOS, Landmark wordnet data & 77.35 \% & 52.09\% \\
WOS, Rewrite Data & 75.46\% & 50.72\% \\
WOS, RAG Augment Data & 77.67\% & 52.67\% \\
WOS, Combined Data & \textbf{82.43}\% & \textbf{60.02}\% \\
WOS, Original Data & 90.32\% & 81.01\% \\
\bottomrule
\end{tabular}
\caption{Detailed results for WOS dataset}
\label{tab:wos_detailed}
\end{table}
     
From the results presented in Table \ref{tab:wos_detailed} for the WoS dataset, we observe a less pronounced performance improvement for RAG augmentation compared to the Reuters dataset. Specifically, even with the RAG Augmented Data method, the performance in terms of Domain Match and Area Match only marginally exceeds that of the Landmark WordNet data. This aligns with our clustering results, which show that the Homogeneity score is lower when the cluster numbers are the same, and for the Silhouette Score to reach its peak, a higher percentage of data is required. This can be attributed to several factors: research topics are inherently more complex and specialized than news articles, which may render the LLM rewriting and RAG generation processes less effective compared to traditional WordNet methods. Traditional methods typically involve replacing only certain "safe" words, which are more straightforward and less error-prone. Despite these challenges, fine-tuned models consistently outperform the RAG-CoT approach without parameter tuning, highlights the importance of task-specific model fine-tuning. Moreover, it is clear that using all original training data, which includes over 20,000 labeled samples, leads to significantly better overall performance. However, in real-world scenarios where labeled data is often limited, our methods for generating additional augmented data still contribute to huge boost in performance improvements. This emphasizes the significance of data augmentation techniques in enhancing model performance, especially when sample sizes are constrained. Notably, combining the Landmark, rewrite, and RAG Augment data results in a substantial improvement, with an 8.58\% increase in Domain Match and a 16.33\% increase in Area Match compared to using only the landmark data. The significant improvement in area classification is particularly encouraging, as it suggests that the different levels of vocabulary overlap in the combined augmented data cover a broader range of case types, thereby leading to a significant performance boost when used together rather than individually. These findings not only emphasize the critical role of data augmentation in boosting model performance but also suggest that different strategies should be adopted for optimizing model performance across various types of text data.\\

\section{Ablation: Random compared with Landmark data for LLMs}

In addition to checking the effectiveness of the augmentation method, we evaluated our clustering techniques' performance compared to random selection through a comparative analysis. This analysis involved two main types of data: landmark data and random data. The landmark data were carefully selected to represent meaningful and representative samples from the dataset, reflecting core themes. In contrast, the random data were chosen without specific criteria, serving as a baseline to gauge the added value of the clustering technique. To ensure a fair assessment of randomness, we conducted three trials for the random data due to computational constraints:

\begin{itemize}
    \item \textbf{Random Data}: An equivalent number of samples (300 from Reuters and 600 from WoS) randomly chosen from the training set to serve as a control group. 
    \item \textbf{Random-WordNet Data}: Similarly, the random data was also augmented tenfold using WordNet, resulting in 3,000 samples from Reuters and 6,000 from WoS. 
\end{itemize}

\begin{table}[ht]
\centering
\begin{tabular}{llll}
\toprule
\textbf{Dataset} & \textbf{Part Match} & \textbf{All Match} & \textbf{In Right Order}  \\ 
\midrule
Trail 1, random data & 88.92\% & 80.81\% & 74.28\% \\
Trail 2, random data & 88.13\% & 82.90\% & 76.08\% \\
Trail 3, random data & 87.85\% & 80.76\% & 74.23\% \\
Landmark data & 90.85\% & 82.35\% & 75.86\% \\
Trail 1, random wordnet data & 89.52\% & 82.25\% & 76.43\% \\
Trail 2, random wordnet data & 89.57\% & 83.45\% & 75.57\% \\
Trail 3, random wordnet data & 88.13\% & 81.74\% & 73.54\% \\
Landmark wordnet data & \textbf{93.05\%}  & \textbf{83.97\%} & \textbf{78.82\%} \\
\bottomrule
\end{tabular}
\caption{Performance Comparison: Landmark vs. Random Data for Reuters.}
\label{tab:reuters_random_comparison}
\end{table}

The results, as summarized in Table \ref{tab:reuters_random_comparison} for the Reuters dataset, reveal several interesting insights. Initially, it might seem counterintuitive that in some trials, random data outperformed landmark data in terms of "All match" and "In right order" accuracy. However, this phenomenon can be attributed to the fact that when the data volume is small, more data about the majority classes is needed to improve model performance. During random selection, more samples are likely to fall into the majority class. Additionally, the test set contains more data representing common cases rather than edge cases. Therefore, in the initial stage of improving accuracy, more data describing the majority of the dataset is required to ensure the model generalizes well to typical scenarios. Despite this initial finding, the true value of the landmark data becomes evident when the datasets are augmented with WordNet. The simple augmentation process generated more samples for each category, which in turn significantly boosted the performance of the landmark data over the random data. After augmentation, the Part Match metric for Reuters landmark data rose from 90.85\% to 93.05\%, a 2.2\% improvement compared to a 0.77\% boost with random data. Similarly, All Match increased by 1.62\% (from 82.35\% to 83.97\%), and In Right Order by 2.95\% (from 75.86\% to 78.82\%), significantly outpacing random data improvements of 0.99\% and 0.31\%, respectively. These results indicate that the clustering technique used to select the landmark data provides more meaningful and valuable insights for enhancing model performance.\\

\begin{table}[ht]
\centering
\begin{tabular}{lll}
\toprule
& \textbf{Domain Match} & \textbf{Area Match} \\ 
\midrule
Trail 1, random data & 74.84\% & 47.13\% \\
Trail 2, random data & 77.25\% & 47.09\% \\
Trail 3, random data & 72.03\% & 46.10\% \\
Landmark data & 73.85\% & 47.48\% \\
Trail 1, random wordnet data & 76.99\% & 51.21\% \\
Trail 2, random wordnet data & 77.32\% & 50.57\% \\
Trail 3, random wordnet data & 77.30\% & 50.06\% \\
Landmark wordnet data & \textbf{77.35}\% & \textbf{52.09}\% \\
\bottomrule
\end{tabular}
\caption{Performance Comparison: Landmark vs. Random Data for WOS.}
\label{tab:wos_random_comparison}
\end{table}

Similarly, for the results shown in Table \ref{tab:wos_random_comparison}, a significant performance improvement was observed for both landmark and random data in Domain Match and Area Match after applying WordNet augmentation. Notably, the landmark-WordNet data achieved the highest accuracy in both Domain Match (77.35\%) and Area Match (52.09\%), representing increases of 3.5\% and 4.61\%, respectively. The significant improvement in Area Match for the landmark-WordNet data is particularly noteworthy, as Area Match is a more granular metric that requires higher precision from the model. Without augmentation, the landmark data slightly underperformed compared to the random data in both metrics. Similar to the Reuters dataset, this may be because the landmark data selection focuses on capturing core themes within the dataset, which may not always align perfectly with the overall distribution of the majority of samples. In contrast, random data selection tends to better reflect the overall distribution of the dataset, especially when the sample size is small, making it more likely to cover the main categories.\\

\section{Ablation: Test the performance of Augmentation Using Machine Learning Model}

In addition to evaluating the improvements afforded by LLMs, we investigated enhancements achieved through traditional machine learning models. Specifically, we employed three machine learning algorithms: Gradient Boosting Machine (GBM), LightGBM, and Random Forest, all built upon BGE-m3 embeddings of the documents. We then applied an ensemble approach to combine the outputs of these traditional models when applied to the embedded data, to determine the final category assignments. For the categorization of Reuters Partial Match documents, we focused on the top two categories identified by our model, as fine-tuned LLMs predict an average of 1.14 categories per document, fewer than two categories.\\

\begin{table}[ht]
\centering
\begin{tabular}{lccccc}
\toprule
& \multicolumn{2}{c}{\textbf{Reuters}} & \multicolumn{2}{c}{\textbf{WOS}} \\
\cmidrule(lr){2-3} \cmidrule(lr){4-5}
& \textbf{Part Match} & \textbf{All Match} & \textbf{Domain Match} & \textbf{Area Match} \\
\midrule
Landmark Data & 78.54\% & 70.06\% & 66.97\% & 19.03\% \\
WordNet Data & 78.04\% & 68.03\% & 65.97\% & 24.26\% \\
LLM Rewrite & 77.15\% & 67.56\% & 66.74\% & 25.72\% \\
RAG Augment & 88.46\% & \textbf{79.65\%} & \textbf{73.60\%} & 39.46\% \\
Combine Data & \textbf{89.11\%} & 79.61\% & 73.91\% & \textbf{40.60\%} \\
\bottomrule
\end{tabular}
\caption{Performance Comparison of the Machine Learning Ensemble Model with Various Data Augmentation Techniques.}
\label{tab:traditional_comparison_methods}
\end{table}

As illustrated in Table~\ref{tab:traditional_comparison_methods}, when working with traditional machine learning models, data augmentation methods that result in a high level of embedding similarity can actually degrade model performance. Conversely, the RAG augmentation method, which generates diverse variations of the data while preserving semantic coherence, significantly boosts performance. The RAG method achieves the highest accuracy for the Reuters All Match and WOS Domain datasets, underscoring its effectiveness in enhancing traditional machine learning models.

\section{Conclusion and Future Works}

In this study, we introduce an innovative framework for fine-tuning large language models (LLMs) in text classification tasks with few labeled data points. This framework integrates several techniques, including clustering for data selection, retrieval-augmented generation (RAG), rewriting using LLMs, and word replacement using WordNet for data augmentation. Our experimental results show that employing clustering algorithms for data selection effectively identifies representative landmark examples and finds near neighbors, thereby enhancing the RAG process and significantly improving the model's accuracy over random selection and other existing methods. This improvement is especially noticeable in datasets like Reuters 20 Newsgroups and Web of Science, which involve categorizing documents into over 100 classes. Our novel data augmentation approach, leveraging RAG and clustering, not only expands the dataset but also enhances its diversity and richness, leading to substantial improvements in classification accuracy. By incorporating all stages of augmented data—from LLM rewrites through RAG to traditional word replacements with varying degrees of vocabulary overlap—our approach ensures that the training data remains diverse and representative, thus enabling the model to better handle various practical scenarios and improving its generalization and accuracy. Furthermore, our methods, which rely on open-source tools, are straightforward to implement, making them highly practical and accessible. Overall, our study provides a robust and efficient solution for text classification tasks, offering significant potential for widespread adoption and application. For future research, several areas can be explored to further improve our framework. One potential direction is to implement constrained decoding so that LLMs generate outputs limited to the available categories and conform to specified regular expressions This can be crucial for achieving 100\% usability in practical applications. Another area is to explore different methods for data augmentation and compare them with our current approach. Although we utilized LLMs for selecting landmarks, alternative methods based on numerical metrics could potentially prove more effective. Additionally, scaling up the model to larger LLMs and comparing their performance against even larger models, like the QLORA 70B, would offer valuable insights into the scalability and efficiency of our framework.

\bibliographystyle{unsrt}  
\bibliography{references}  

\appendix
\setcounter{section}{0} 

\section{Appendix, Detail For Selecting Clustering Methods}\label{appendix:cluster}

\subsection{Text Preprocessing}

In tasks involving document text processing, a standard preliminary step involves tokenization, where the document is broken down into a sequence of individual words. Following this, a transformation process is commonly applied to convert the textual data into numerical representations to enable further analysis, such as clustering and classification. Techniques like Term Frequency-Inverse Document Frequency (TF-IDF)\cite{5c392ec2df5b8c0b3c887796} and fixed-length embeddings\cite{53e9acc4b7602d97036a1037} are predominant in this conversion. Embedding methodologies transform document words into a series of vector representations, which can be seamlessly integrated into Large Language Models (LLMs) for downstream tasks. This direct translation facilitates a deeper understanding of semantic relationships within the text. Conversely, frequency-based methods like TF-IDF offer a complementary perspective from a statistical summary viewpoint, both within and across documents, and typically operate much faster for large text datasets.

\subsection{Clustering}

From the standard K-means clustering to those enabling the automatic adjustment of cluster center numbers, clustering has long been a traditional unsupervised learning method. LMSC\cite{5c0f8eabda562944aca3fe11} utilized the complementary information from multiple views of the data. It reconstructed original data by multiplying different projection vectors and subsequently performed clustering on these subspaces. When the number of available documents is large, it is usually time-consuming to conduct all clustering simultaneously. LBPSO\cite{5ac1827b17c44a1fda915546} adopts a neighborhood selection strategy, learning from individual historical best solutions and neighborhood best solutions, avoiding calculating a global solution at once and preventing the algorithm from getting entraped in local optima. \cite{5ff75c42d4150a363c6f3e84} mentioned a local fuzzy one pass clustering method where the one pass stands for accomplishing the clustering process in a single iteration over the dataset, thus can handle streaming incremental data. 

\subsection{NMI}

Mutual Information (MI) quantifies the amount of information one clustering provides about another, making it a useful measure of similarity between two clusterings of the same dataset. Given two clusterings \(U\) and \(V\), MI is defined as:

\[ I(U; V) = \sum_{u \in U, v \in V} \frac{|u \cap v|}{N} \log \left( \frac{N \cdot |u \cap v|}{|u| \cdot |v|} \right) \]

To address the potential bias due to cluster size differences and random cluster alignments, Normalized Mutual Information (NMI) standardizes MI. NMI scales MI by the average entropy of the two clusterings, ensuring the measure ranges from 0 to 1:

\[ \text{NMI}(U; V) = \frac{I(U; V)}{\text{avg}(\text{H}(U), \text{H}(V))} \]

Here, \(\text{H}(X)\) is the Shannon entropy of clustering \(X\). An NMI of 1 indicates perfect agreement, while values close to 0 suggest near-independence, making NMI a valuable tool for evaluating clustering algorithms.

\subsection{Selection of Clustering methods}

There are a variety of clustering techniques and feature representation methods in our disposal. For the data selection phase, we selected four clustering algorithms and two feature representation strategies that are particularly well-suited for scenarios with large numbers of samples and clusters. The chosen clustering methods include Bisecting K-Means, Balanced Iterative Reducing and Clustering Using Hierarchies (BIRCH)\cite{53e9b234b7602d9703cd29a4}, Hierarchical Clustering and GaussianMixture Models. For feature representation, BGE-m3 Embedding\cite{65c19b55939a5f40825f9ad2} and TF-IDF techniques were chosen. The BGE-m3 method embeds up to 8192 tokens into 1024-dimensional vectors. For TF-IDF, we selected severa different max features terms based on frequency scores to address sparsity in the TF-IDF matrix. To evaluate the suitability of these combinations for our specific needs, we measure their performance using several evaluation metrics. Given that we possess the ground truth labels and our goal is to use clustering to identify more representative samples rather than just assigning predicted cluster labels, we are particularly focused on the Homogeneity score, which measures the extent to which each cluster contains only members of a single class. In addition to homogeneity, we also employ the Normalized Mutual Information (NMI) and Silhouette Score to assess the quality of the clustering.

\begin{table}[h]
\centering
\caption{Comparison of Embedding Methods with TF-IDF Methods under Different Max-Features Settings for Reuters-21578 Dataset, using Hierarchical Clustering Methods, under the constrain of 300 clusters.}
\begin{tabular}{cccc}
\toprule
\multicolumn{1}{c}{\textbf{Tf-idf Max Features}} &
\textbf{Homogeneity} &
\textbf{NMI} &
\textbf{Silhouette Score} \\
\midrule
256, TF-IDF                  & 0.7878            & 0.4784                &  \textbf{0.0726}       \\
512, TF-idf                  & 0.7929            & 0.4832                & 0.0723       \\
1024, TF-idf                 & 0.8104            & 0.4936                & 0.0657       \\
2048, TF-idf                 & 0.8003            & 0.4916                & 0.0641       \\
4096, TF-idf                 & 0.7895            & 0.4875                & 0.0596       \\
-, Embedding                 & \textbf{0.8338}            & \textbf{0.5123}                & 0.0579       \\
\bottomrule
\end{tabular}
\label{cluster:table1}
\end{table}

To evaluate the effectiveness of different feature extraction methods (TF-IDF and Embedding) on text clustering, we conducted experiments using the Reuters-21578 dataset with varying dimensions of TF-IDf max feature spaces (256, 512, 1024, 2048, 4096) and Hierarchical clustering algorithm, under the constrain of 300 clusters. Table~\ref{cluster:table1} summarizes the results in terms of homogeneity, Normalized Mutual Information (NMI), and Silhouette Score. For TF-IDF based clustering on the Reuters-21578 dataset, the optimal homogeneity and NMI are achieved with 1024 max features. Deviating from this setting slightly reduces these metrics. The Silhouette Score consistently declines as the number of features decreases. Embeddings outperform TF-IDF with hierarchical clustering, achieving the highest Homogeneity (0.8338) and NMI (0.5123). Additionally, embedding methods do not require tuning for max features, unlike TF-IDF methods. Other clustering methods yield similar results.

\begin{table}[h]
\centering
\caption{Comparison of Embedding Methods with TF-idf Methods under Different Max-Features Settings for WOS-46985 Dataset, using Hierarchical Clustering Methods, under the constrain of 600 clusters.}
\begin{tabular}{cccc}
\toprule
\multicolumn{1}{c}{\textbf{Tf-idf Max Features}} &
\textbf{Homogeneity} &
\textbf{NMI} &
\textbf{Silhouette Score} \\
\midrule
512, TF-idf                  & 0.5058            & 0.4432                & 0.0226       \\
1024, TF-idf                 & 0.5455            & 0.4779                & \textbf{0.0306}       \\
2048, TF-idf                 & 0.5573            & 0.4907                & 0.0277       \\
4096, TF-idf                 & 0.5547            & 0.4925                & 0.0212       \\
8192, TF-idf                 & 0.5453            & 0.4879                & 0.0136       \\
-, Embedding          & \textbf{0.6530} & \textbf{0.5677} & 0.0116  \\ \hline
\bottomrule
\end{tabular}
\label{cluster:table2}
\end{table}

As shown in Table~\ref{cluster:table2}, embedding methods consistently outperform TF-IDF methods across all different TF-IDF maximum feature settings for the WOS-46985 dataset. This suggests that embedding methods are better at capturing the deep semantic structures within the text, which is crucial for the classification criteria of the WOS-46985 dataset. Furthermore, the optimal vocabulary size for TF-idf was found to be between 2048 and 4096, indicating that scientific documents require a larger vocabulary for accurate representation compared to Reuters news articles. The lower Homogeneity and NMI scores for TF-idf also suggest a more challenging classification problem. Detailed grid search results confirmed that embedding methods achieve higher homogeneity and NMI scores, leading us to adopt embedding methods for our clustering approach.

From Tables \ref{cluster:table3} and \ref{cluster:table4}, it can be observed that Hierarchical Clustering and the Gaussian Mixture Model (GMM) outperform other clustering methods in most scenarios. While BIRCH and Bisecting K-Means are algorithms designed for computational efficiency, Hierarchical Clustering and GMM require more processing time, with GMM being the most computationally demanding. In practice, the output of BIRCH is often used as an initial point for hierarchical clustering or GMM. In addition to the encoding method employed, the size of the cluster numbers also impacts the homogeneity and normalized mutual information (NMI) scores. As the number of clusters increases, both homogeneity and NMI scores tend to rise. We also observed that the Gaussian Mixture Model (GMM) exhibits better performance compared to the Hierarchical Model when the number of clusters is smaller. However, as the size of the clusters increases, the Hierarchical Model surpasses GMM in turns of NMI, Homogeneity and Silhouette Score. An additional advantage of the Gaussian Mixture Model is its probabilistic nature, which enhances the interpretability of cluster membership. GMM provides probabilities for each data point belonging to different clusters, thereby allowing us to identify multiple potential cluster affiliations for each data point. This feature makes GMM particularly suitable for our subsequent step involving rag retrieval. It's important to note that both Gaussian Mixture Models (GMM) and hierarchical clustering can generate a list of the most likely clusters for each sample. GMM provides a probabilistic framework, assigning a probability density to each sample across all clusters, whereas hierarchical clustering offers a nested structure of subclusters within larger clusters, which can be used to identify the nearest clusters for a given sample.

\begin{table}[h]
\centering
\caption{Comparison of Clustering Methods with Different Cluster Sizes on the Reuters Dataset}
\label{cluster:table3}
\adjustbox{width=0.60\linewidth}{
\begin{tabular}{@{}llcccc@{}}
\toprule
\textbf{Clustering Method} & \textbf{Number of Clusters} & \textbf{Homogeneity} & \textbf{NMI} & \textbf{Silhouette Score} \\ \midrule
Random                     & 100                         & 0.1195               & 0.0720       & -0.0239                         \\
Random                     & 300                         & 0.2679               & 0.1389       & -0.0513                          \\
Random                     & 600                         & 0.4042               & 0.1934       & -0.1037                          \\ \midrule
Brich                      & 100                         & 0.7299               & 0.5194       & 0.0354                     \\
Brich                      & 300                         & 0.8147               & 0.5119       & 0.0446                     \\
Brich                      & 600                         & 0.8593               & 0.5021       & 0.0554                     \\ \midrule
Bisecting K-Means          & 100                         & 0.7101               & 0.4941       & 0.0230                     \\
Bisecting K-Means          & 300                         & 0.7916               & 0.4777       & 0.0127                     \\
Bisecting K-Means          & 600                         & 0.8473               & 0.4702       & 0.0034                     \\ \midrule
Hierarchical               & 100                         & 0.7467               & 0.5335       & 0.0421                     \\
Hierarchical               & 300                         & 0.8338               & 0.5123       & 0.0579                     \\
Hierarchical               & 600                         & 0.8839               & 0.4963       & 0.0835                     \\ \midrule
Gaussian Mixture            & 100                         & 0.7513               & 0.5275       & 0.0389                     \\
Gaussian Mixture            & 300                         & 0.8127               & 0.4934       & 0.0408                     \\
Gaussian Mixture            & 600                         & 0.8765               & 0.4944       & 0.0583                     \\ \bottomrule
\end{tabular}}
\end{table}

\begin{table}[h]
\centering
\caption{Comparison of Clustering Methods with Different Cluster Sizes for the WOS Dataset}
\label{cluster:table4}
\adjustbox{width=0.60\linewidth}{
\begin{tabular}{@{}lcccc@{}}
\toprule
\textbf{Clustering Method} & \textbf{Number of Clusters} & \textbf{Homogeneity} & \textbf{NMI} & \textbf{Silhouette Score} \\ \midrule
Random           & 150   & 0.1565 & 0.1539 & -0.0109  \\
Random           & 300   & 0.2551 & 0.2345 & -0.0171  \\
Random           & 600   & 0.3713 & 0.3206 & -0.0259  \\
Random           & 1200 & 0.4962 & 0.4044 & -0.0424  \\ \midrule
Birch            & 150   & 0.5141 & 0.5099 & 0.0067  \\
Birch            & 300   & 0.5785 & 0.5369 & 0.0087  \\
Birch            & 600   & 0.6420 & 0.5609 & 0.0097  \\
Birch            & 1200  & 0.7087 & 0.5859 & 0.0138  \\ \midrule
Bisecting K-Means& 150   & 0.4738 & 0.4685 & 0.0051  \\
Bisecting K-Means& 300   & 0.5338 & 0.4943 & 0.0002  \\
Bisecting K-Means& 600   & 0.5940 & 0.5168 & -0.0077 \\
Bisecting K-Means& 1200  & 0.6642 & 0.5436 & -0.0176 \\ \midrule
Hierarchical     & 150   & 0.5240 & 0.5202 & 0.0073  \\
Hierarchical     & 300   & 0.5865 & 0.5440 & 0.0081  \\
Hierarchical     & 600   & 0.6530 & 0.5677 & 0.0116  \\
Hierarchical     & 1200  & 0.7225 & 0.5917 & 0.0190  \\ \midrule
Gaussian Mixture & 150   & 0.5516 & 0.5452 & 0.0206  \\
Gaussian Mixture & 300   & 0.6017 & 0.5564 & 0.0177  \\
Gaussian Mixture & 600   & 0.6532 & 0.5681 & 0.0130  \\
Gaussian Mixture & 1200  & 0.7033 & 0.5774 & 0.0092  \\ \bottomrule
\end{tabular}}
\end{table}

\section{Analysis of Generated data}\label{sec:gen_data_desc}

To refine the selection of representative samples from each cluster, we implemented a multi-step process. Initially, all samples within each cluster were indexed. Subsequently, LLMs were employed to identify and select the sample that most accurately embodies the core characteristics of its respective cluster. This was achieved by instructing the LLMs to return the index corresponding to the most representative sample, as illustrated in \hyperlink{tcb:repchoose1}{Table C.5}. Prior to this analysis, the documents underwent preprocessing, which involved stripping line spaces—replacing newline characters with spaces—and compressing multiple spaces into single ones. Following this preprocessing, the longest prompts for the Reuters and Web of Science (WoS) datasets contained 14,294 and 21,899 words, respectively, both comfortably within the acceptable length limits for our models. The LLMs' responses, typically provided in the form of a numerical index, were extracted using a regular expression pattern `[$\backslash$d+]`, and the last matched answer was utilized for further processing.

Following the selection and revelation of the true labels of the most representative landmarks from each cluster, we proceeded to utilize all available unlabeled data, along with the labeled landmarks (\(X\), \(X_{\pi}^{m}\), \(Y^m\)), to generate additional training data. For the RAG generation using prompt as shown in \hyperlink{tcb:augment1}{Table C.4}, the average prompt length was 1,836.25 words for the Reuters dataset and 2,260.34 words for the Web of Science (WoS) dataset, with the longest prompts comprising 5,607 and 3,573 words, respectively. For each document, we utilized the prompt to generate three unique samples by varying the randomly selected reference documents. Subsequently, we applied regular expressions to parse out the relevant context and labels from the generated texts. Out of the 16,182 documents generated for the Reuters dataset, 15,473 successfully underwent the regular expression-based extraction of context and labels. Similarly, from the 70,476 documents generated for the WoS dataset, 62,886 documents passed through the extraction process. Additionally, for the rewriting task using large LLMs, as detailed in \hyperlink{tcb:rewrite1}{Table C.3}, we generated ten samples for each of the landmark documents within the datasets. This was conducted with the hyperparameter temperature set to 0.3, yielding a total of 3,000 rewritten documents for the Reuters dataset and 6,000 for the WoS dataset. Since these generated samples did not require label extraction, they all successfully passed the regular expression filtering phase. It is noteworthy that the prompt lengths for the rewriting tasks were significantly shorter compared to those used for RAG generation, averaging 254.95 words for the Reuters dataset and 251.4 words for the WoS dataset.

\begin{figure}[h]
\centering
\includegraphics[width=0.49\textwidth]{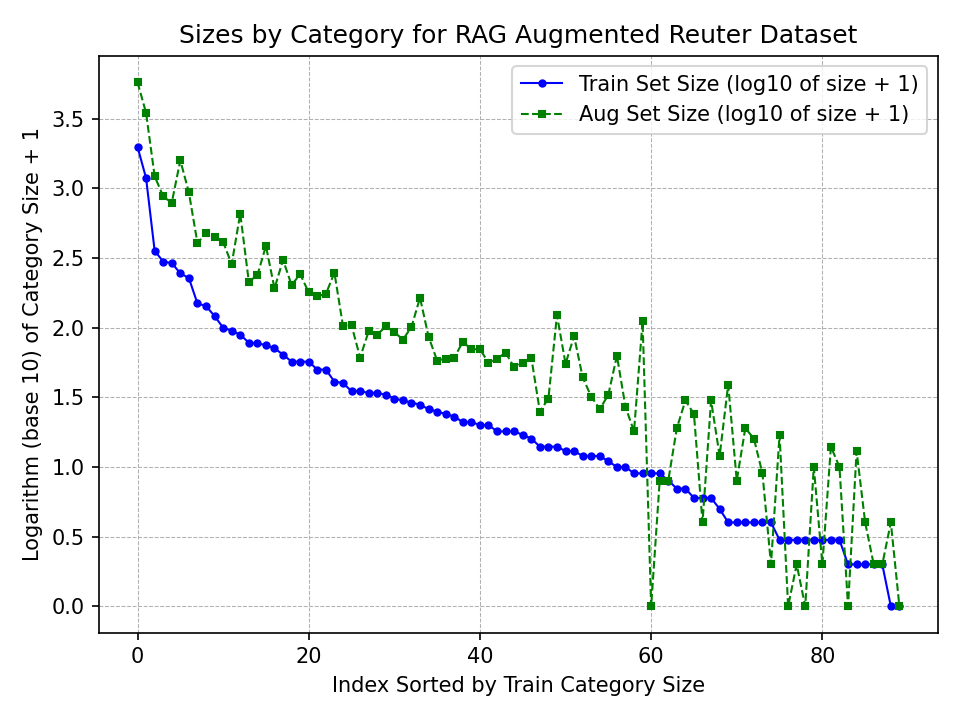}
\includegraphics[width=0.49\textwidth]{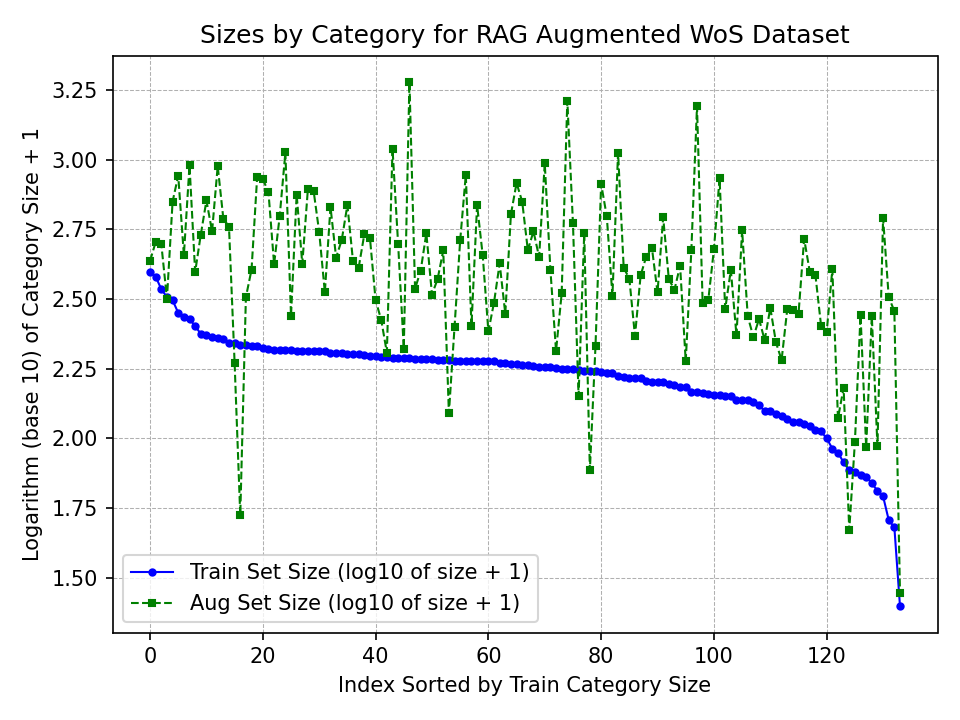}
\caption{Comparison of Category Distributions Before and After Augmentation with Three Samples per Document in the Training Set.}
\label{fig:category}
\end{figure}

For the Reuters dataset, we configured the LLMs to produce multiple labels for each document, with individual labels delineated by commas. In contrast, for the WoS dataset, the LLM was instructed to generate a hierarchical labeling structure, beginning with the domain and then specifying the sub-domain, with each level separated by commas. Subsequently, the generated documents were evaluated against their original counterparts using a suite of performance metrics. While the labels for LLMs rewrite are kept the same as the original labels, we investigated the labeling of RAG-generated examples, as depicted in Figure~\ref{fig:category}. The Reuters dataset originally comprised 90 distinct labels, which were expanded to 443 unique labels through the use of LLMs. This expansion partly results from LLMs generating hallucinatory labels due to lack of fine-tuning for precise label generation. This process utilized 300 labeled landmarks, with 59 labels being covered. Despite this, 93.49\% of documents in the augmented dataset still carry labels from the original 90, covering 85 of them. The training set includes 87 out of the original 90 labels.

While the labels for LLMs rewrite are kept the same as the original labels, we investigated the labeling of RAG-generated examples, as depicted in Figure~\ref{fig:category}. The Reuters dataset originally comprised 90 distinct labels, which were expanded to 443 unique labels through the use of LLMs. This expansion partly results from LLMs generating hallucinatory labels due to lack of fine-tuning for precise label generation. This process utilized 300 labeled landmarks, with 59 labels being covered. Despite this, 93.49\% of documents in the augmented dataset still carry labels from the original 90, covering 85 of them. The training set includes 87 out of the original 90 labels. Compared to the original dataset, the augmented version shows increased category sizes. While only 21 categories had over 50 samples originally, now 52 categories exceed this threshold. It is expected that this number will increase further as more samples are augmented. However, there are 9 instances where the number of augmented documents is fewer than the original, often due to rare terms such as "jet," "cotton-oil," "dfl," "cpu," and "naphtha," which appeared only one or two times in the training data. Specifically, the "lei" category, representing the Leading Economic Index, saw no augmentation despite having 8 instances in the training set, likely because LLMs failed to recognize "lei" without context. For the Web of Science (WoS) dataset which comprising 7 domains and 134 sub-areas, the augmented dataset introduces 11 domains, with 99.90\% of the new domains aligned with the original 7. Regarding sub-areas, the augmented dataset generates 2,765 new sub-areas, of which 82.63\% remain within the 134 sub-areas of the original WoS dataset, and 133 of the 134 original sub-areas are included. There are 7 cases where the augmented data has fewer samples compared to the original data. Upon examination, the least number of augmented documents, "medical, polycythemia vera" consists of 46, compared to 76 in the training set. The proportionate least is "psychology, attention," with 216 in the training set and only 52 in the augmented data. 

\begin{figure}[h]
\centering
\includegraphics[width=0.49\textwidth]{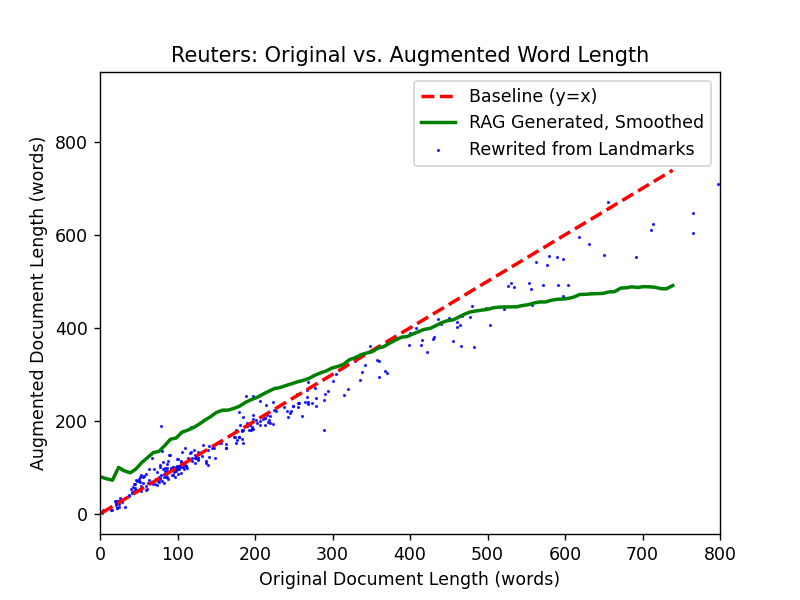}
\includegraphics[width=0.49\textwidth]{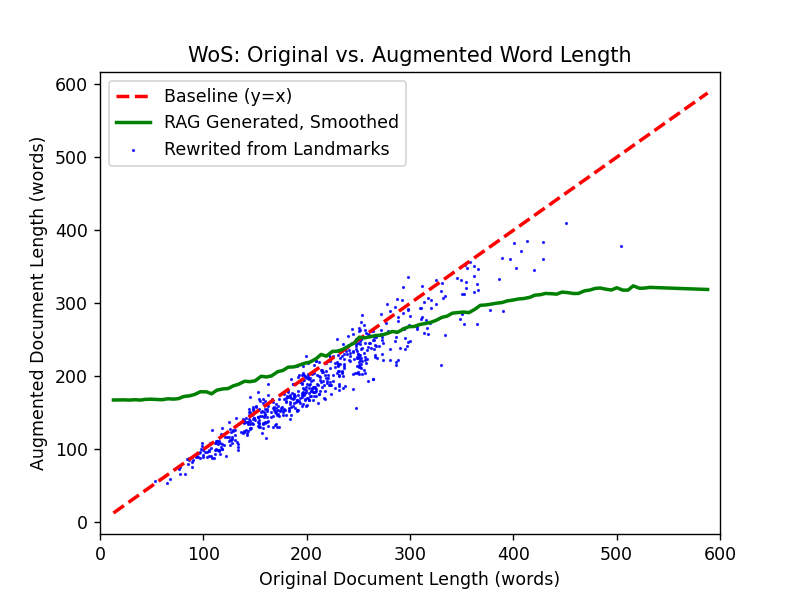}
\caption{Comparison of Smoothed Word Lengths per Document Between Original and Augmented Documents in the Reuters and WoS Datasets}
\label{fig:gen_length}
\end{figure}

The average word count for the original context in the Reuters dataset is 128.5 for the training set, whereas the augmented data exhibits an average length of 167.4 words for RAG augmentation. For rewrite augmentation, the bases for augmentation are the landmarks, average length have 177.39 and 177 words for the augmentation.  Regarding the WoS dataset, the initial documents within the training set have an average word count of 198.3, while the augmented documents extend to an average of 218.3 words. For rewrite augmentation, the bases for augmentation are the landmarks, average length have 208.40 and 191.24 words for the augmentation. It is observed that for RAG augmentation, when the original documents were relatively short, the synthesized versions tended to be more expansive. Conversely, if the original documents were already quite lengthy, the synthesized documents only showed a marginal increase in length on average. While for LLM rewrite, the generated documents are on average a little bit shorter than original. These observations are visually depicted in Figure~\ref{fig:gen_length}. When considering the average character length per word, there is a slight increase of 1.4\% for the Reuters dataset and 2.3\% for the WoS dataset; these increments are statistically significant, as indicated by a very small standard error. A small indication that LLMs tends to use complex words.\\

\begin{figure}[h]
\centering
\includegraphics[width=1\textwidth]{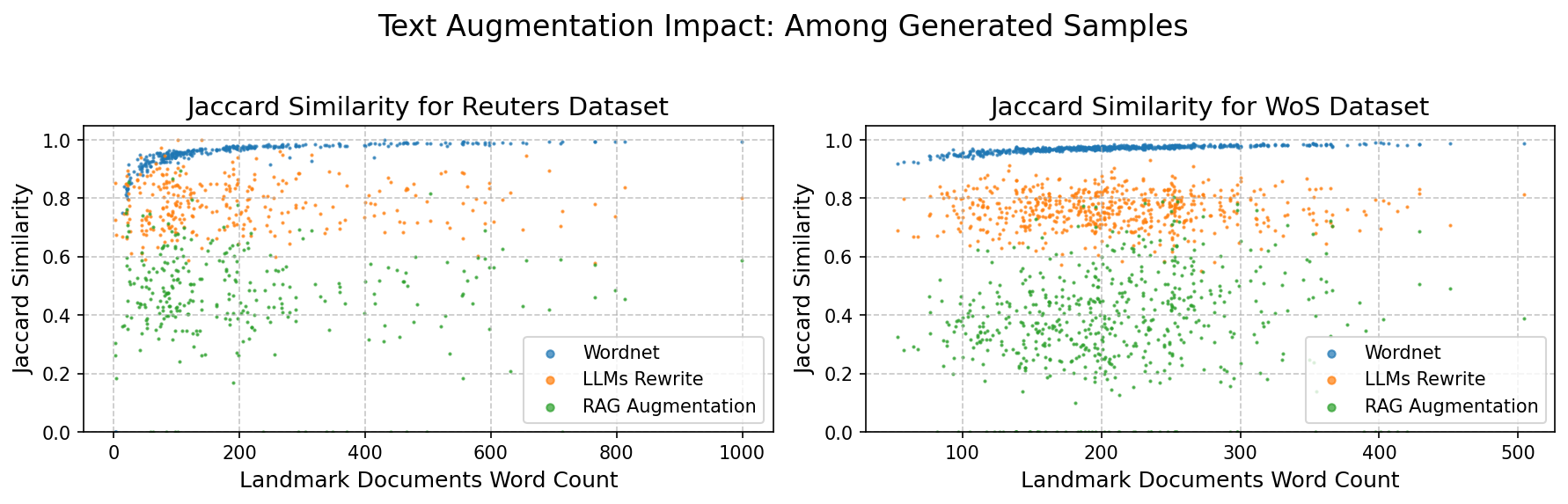}
\caption{Impact of Text Augmentation on Jaccard Similarity (Vocabulary Overlap) among Generated Samples: The figure illustrates that from RAG augmentation to LLMs rewrite to traditional WordNet random synonym replacement, the vocabulary overlap decreases, indicating the generation of more diverse samples.}
\label{fig:gen_jaccard}
\end{figure}

\section{Appendix: Prompts for Generation}

\noindent 
\begin{minipage}[b]{0.49\linewidth}
\begin{tcolorbox}[colback=white, title = \hypertarget{tcb:generate1}{} \textbf{Table C.1:} Prompt For LLM Finetuning, colframe=black, sharp corners, width=\linewidth]
Assign tags for the following \{subject\} Document:\\

\{Document\}\\

Answer:

[label1, label2, ... ]
\end{tcolorbox}
\end{minipage}%
\hfill 
\begin{minipage}[b]{0.49\linewidth}
\begin{tcolorbox}[colback=white, title = \hypertarget{tcb:generate2}{} \textbf{Table C.2:} Prompt For Label Prediction, colframe=black, sharp corners, width=\linewidth]
Assign tags for the following \{subject\} Document:\\

\{Document\}\\

Answer:
\\
\end{tcolorbox}
\end{minipage}

\begin{tcolorbox}[colback=white, title =  \hypertarget{tcb:rewrite1}{ \textbf{Table C.3:} } Prompt For LLMs Document Rewrite,colframe=black, sharp corners, label=tcb:rewrite1]
*Task Description: Rewrite the following text in English, maintaing the original meaning but using different words and sentence structures. The new version should be clear and concise, and it should not alter the core message of the original text.\\

*Original Text:\\
\{The document upon which augmentation will be based\}\\

Rewritten Text:
\end{tcolorbox}

\begin{tcolorbox}[colback=white, title =\hypertarget{tcb:augment1}{\textbf{Table C.4:}} Prompt For RAG Dataset Augmentation,colframe=black, sharp corners, label=tcb:augment1]
*Task Description:
You are provided with a set of similar documents, some of which are labeled and others are not. Your task is to generate a sample document based on the primary document, using both the labeled and unlabeled documents as references.\\

*List of  Available Labels:\\
\{List of available labels, seperated by comma\}\\

*Reference Labeled Documents:\\
\{A set of labeled landmarks from the nearest five clusters\}\\

*Reference Unlabeled Documents:\\
\{A set of unlabeled documents within the same cluster\}\\

*Primary Document for Augmentation:\\
\{The document upon which augmentation will be based\}\\

*Task:\\
Using the labeled and unlabeled documents as a guide, create a new document based on the primary document and assign it the appropriate labels from the available list.\\

*Document Format:\\
Content:\\
Label: [Your assigned label]\\

*Generated Example:
\end{tcolorbox}

\begin{tcolorbox}[colback=white, title = \hypertarget{tcb:repchoose1}{\textbf{Table C.5}}: Prompt to Choose Representative Landmarks for Each Cluster ,colframe=black, sharp corners]
You have been provided with a set of similar documents, each indexed by a number. Your task is to identify the most representative example from this cluster of documents. Please carefully analyze the given documents and select one document that best captures the common essence and characteristics of the samples. The selection should emphasize the representativeness and relevance of the chosen sample to the category, so that it can serve as a reference for labeling the entire cluster.\\

Documents in the cluster:\\

1-. \{Document 1\}...... \\

2-. \{Document 2\}......\\
  
3-. \{Document 3\}......\\
  
4-. \{Document 4\}......\\
 
 $\vdots$\\
 
Please choose one document that could best serve as a reference for labeling the entire cluster, and return only the index number of your selection, in format such as [0], [1], etc.\\

Answer:\\
\end{tcolorbox}

\begin{tcolorbox}[colback=white, title = \hypertarget{tcb:labeling1}{ \textbf{Table C.6:} }  Prompt For RAG CoT Document Labeling,colframe=black, sharp corners]
*Task Description: You are provided with a set of similar documents. Your task is to predict the label for the target document, using the labeled document examples as references.\\

*List of  Available Labels:\\
\{List of available labels, seperated by comma\}\\

*Reference Labeled Documents:\\
\{A set of labeled landmarks from the nearest five clusters\}\\
*Target Document for Prediction:\\
\{The document upon which augmentation will be based\}\\

*Task:\\
Predict the label for the target document. Please provide your reasoning before asssigning the label.\\

*Format:
Thought: [Your thoughts]\\
Label: [Your assigned label]\\

Answer:
\end{tcolorbox}

\section{Available Labels for the two Dataset}

\textbf{The list of available labels for WOS dataset:}\\

\textbf{Domain: CS}\\
Area: Algorithm design  , Bioinformatics  , Computer graphics  , Computer programming  , Computer vision  , Cryptography  , Data structures  , Distributed computing  , Image processing  , Machine learning  , Operating systems  , Parallel computing  , Relational databases  , Software engineering  , Structured Storage  , Symbolic computation  , network security

\textbf{Domain: Civil}\\
Area: Ambient Intelligence  , Bamboo as a Building Material  , Construction Management  , Geotextile  , Green Building  , Highway Network System  , Nano Concrete  , Rainwater Harvesting  , Remote Sensing  , Smart Material  , Solar Energy  , Stealth Technology  , Suspension Bridge  , Transparent Concrete  , Underwater Windmill  , Water Pollution

\textbf{Domain: ECE}\\
Area: Analog signal processing  , Control engineering  , Digital control  , Electric motor  , Electrical circuits  , Electrical generator  , Electrical network  , Electricity  , Lorentz force law  , Microcontroller  , Operational amplifier  , PID controller  , Satellite radio  , Signal-flow graph  , Single-phase electric power  , State space representation  , System identification  , Voltage law

\textbf{Domain: MAE}\\
Area: Fluid mechanics  , Hydraulics  , Internal combustion engine  , Machine design  , Manufacturing engineering  , Materials Engineering  , Strength of materials  , Thermodynamics  , computer-aided design

\textbf{Domain: Medical}\\
Area: Addiction  , Allergies  , Alzheimer's Disease  , Ankylosing Spondylitis  , Anxiety  , Asthma  , Atopic Dermatitis  , Atrial Fibrillation  , Autism  , Bipolar Disorder  , Birth Control  , Cancer  , Children's Health  , Crohn's Disease  , Dementia  , Depression  , Diabetes  , Digestive Health  , Emergency Contraception  , Fungal Infection  , HIV/AIDS  , Headache  , Healthy Sleep  , Heart Disease  , Hepatitis C  , Hereditary Angioedema  , Hypothyroidism  , Idiopathic Pulmonary Fibrosis  , Irritable Bowel Syndrome  , Kidney Health  , Low Testosterone  , Lymphoma  , Medicare  , Menopause  , Mental Health  , Migraine  , Multiple Sclerosis  , Myelofibrosis  , Osteoarthritis  , Osteoporosis  , Outdoor Health  , Overactive Bladder  , Parenting  , Parkinson's Disease  , Polycythemia Vera  , Psoriasis  , Psoriatic Arthritis  , Rheumatoid Arthritis  , Schizophrenia  , Senior Health  , Skin Care  , Smoking Cessation  , Sports Injuries  , Sprains and Strains  , Stress Management  , Weight Loss

\textbf{Domain: Psychology}\\
Area: Antisocial personality disorder  , Attention  , Borderline personality disorder  , Child abuse  , Depression  , Eating disorders  , False memories  , Gender roles  , Leadership  , Media violence  , Nonverbal communication  , Person perception  , Prejudice  , Prenatal development  , Problem-solving  , Prosocial behavior  , Schizophrenia  , Seasonal affective disorder  , Social cognition

\textbf{Domain: Biochemistry}\\
Area: Cell biology  , DNA/RNA sequencing  , Enzymology  , Genetics  , Human Metabolism  , Immunology  , Molecular biology  , Northern blotting  , Polymerase chain reaction  , Southern blotting\\

\textbf{The list of available labels for Reuter dataset:}\\
\\
acq, alum, barley, bop, carcass, castor-oil, cocoa, coconut, coconut-oil, coffee, copper, copra-cake, corn, cotton, cotton-oil, cpi, cpu, crude, dfl, dlr, dmk, earn, fuel, gas, gnp, gold, grain, groundnut, groundnut-oil, heat, hog, housing, income, instal-debt, interest, ipi, iron-steel, jet, jobs, l-cattle, lead, lei, livestock, lumber, meal-feed, money-fx, money-supply, naphtha, nat-gas, nickel, nzdlr, oilseed, orange, palladium, palm-oil, pet-chem, platinum, potato, propane, rand, rape-oil, reserves, retail, rice, rubber, ship, silver, soy-oil, soybean, strategic-metal, sugar, sun-oil, tea, tin, trade, veg-oil, wpi, yen, zinc\\
\end{document}